\documentclass[10pt,twocolumn,letterpaper]{article}

\usepackage{iccv}
\usepackage{times}
\usepackage{epsfig}
\usepackage{graphicx}
\usepackage{amsmath}
\usepackage{amssymb}
\usepackage{gensymb}
\usepackage{cuted}
\usepackage{capt-of}
\usepackage{balance}

\clearpage{}\usepackage{xspace}
\usepackage[dvipsnames]{xcolor}
\usepackage{multirow}
\usepackage{ifthen}
\usepackage{algorithmicx}
\usepackage{algorithm}
\usepackage{algpseudocode}
\usepackage{colortbl}  

\usepackage{subcaption}
\usepackage{paralist}
\usepackage{dblfloatfix}
\usepackage{flushend}

\graphicspath{{figures/}}

\newcommand{\faze}[0]{\textsc{Faze}\xspace}

\makeatletter
\newcommand\semitiny{\@setfontsize\notsotiny{6.31415}{7.1828}}
\makeatother

\long\def\ignorethis#1{}

\definecolor{lightgray}{rgb}{0.92,0.92,0.92}
\definecolor{gray}{rgb}{0.35,0.35,0.35}
\definecolor{darkgreen}{rgb}{0.5,0.5,0}
\definecolor{MyBlue}{rgb}{0,0.2,0.8}
\definecolor{MyRed}{rgb}{0.8,0.2,0}
\definecolor{MyGreen}{rgb}{0.0,0.5,0.1}
\definecolor{MyGray}{rgb}{0.4,0.4,0.4}
\definecolor{airforceblue}{rgb}{0.36, 0.54, 0.66}

\newlength\paramargin
\newlength\figmarginstart
\newlength\figmargin
\newlength\subfigmargin
\newlength\secmargin
\newlength\subsecmargin
\newlength\tabmargin
\newlength\tabmarginstart
\newlength\eqmargin
\newlength\algmargin

\usepackage{array}
\newcolumntype{L}[1]{>{\raggedright\let\newline\\\arraybackslash\hspace{0pt}}m{#1}}
\newcolumntype{C}[1]{>{\centering\let\newline\\\arraybackslash\hspace{0pt}}m{#1}}
\newcolumntype{R}[1]{>{\raggedleft\let\newline\\\arraybackslash\hspace{0pt}}m{#1}}

\newcommand{\Paragraph}[1]
{\vspace{1.5mm} \noindent \textbf{#1}}

\def\ie{i.e.,~}
\def\eg{e.g.,~}

\clearpage{}

\usepackage[breaklinks=true,bookmarks=false]{hyperref}

\iccvfinalcopy 

\def\iccvPaperID{1112}

\ificcvfinal\fi
\begin{document}

\title{
Few-Shot Adaptive Gaze Estimation
}

\author{
Seonwook Park\textsuperscript{12*},
Shalini De Mello\textsuperscript{1*},
Pavlo Molchanov\textsuperscript{1},
Umar Iqbal\textsuperscript{1},
Otmar Hilliges\textsuperscript{2},
Jan Kautz\textsuperscript{1}\\
\textsuperscript{1}NVIDIA,\qquad\textsuperscript{2}ETH Z{\"u}rich\\
{\tt\small \{spark,\,otmarh\}@inf.ethz.ch;\quad\{shalinig,\,pmolchanov,\,uiqbal,\,jkautz\}@nvidia.com}
}

\maketitle
\ificcvfinal\thispagestyle{empty}\fi

\newcommand{\customfootnotetext}[2]{{
  \renewcommand{\thefootnote}{#1}
  \footnotetext[0]{#2}}}
\ificcvfinal\customfootnotetext{*}{The first two authors contributed equally.}\fi

\begin{abstract}

Inter-personal anatomical differences limit the accuracy of person-independent gaze estimation networks. Yet there is a need to lower gaze errors further to enable applications requiring higher quality. Further gains can be achieved by personalizing gaze networks, ideally with few calibration samples. However, over-parameterized neural networks are not amenable to learning from few examples as they can quickly over-fit.  
We embrace these challenges and propose a novel framework for Few-shot Adaptive GaZE Estimation (\faze) for learning person-specific gaze networks with very few ($\leq9$) calibration samples. 
\faze learns a rotation-aware latent representation of gaze via a disentangling encoder-decoder architecture along with a highly adaptable gaze estimator trained using meta-learning. It is capable of adapting to any new person to yield significant performance gains with as few as $3$ samples, yielding state-of-the-art performance of $3.18^{\circ}$ on GazeCapture, a 19\% improvement over prior art.
We open-source our code at \url{https://github.com/NVlabs/few_shot_gaze} \footnote{This includes a real-time demo which takes $<$ 10 seconds to record 9 calibration points for a new user and $\sim$ 1 minute to train a personalized network on a laptop with an NVIDIA GTX GeForce 1060 GPU.}.

\end{abstract}

\section{Introduction}

\begin{figure}\centering
\includegraphics[width=1.0\columnwidth,trim={.2cm, 0cm, 0cm, .3cm},clip]{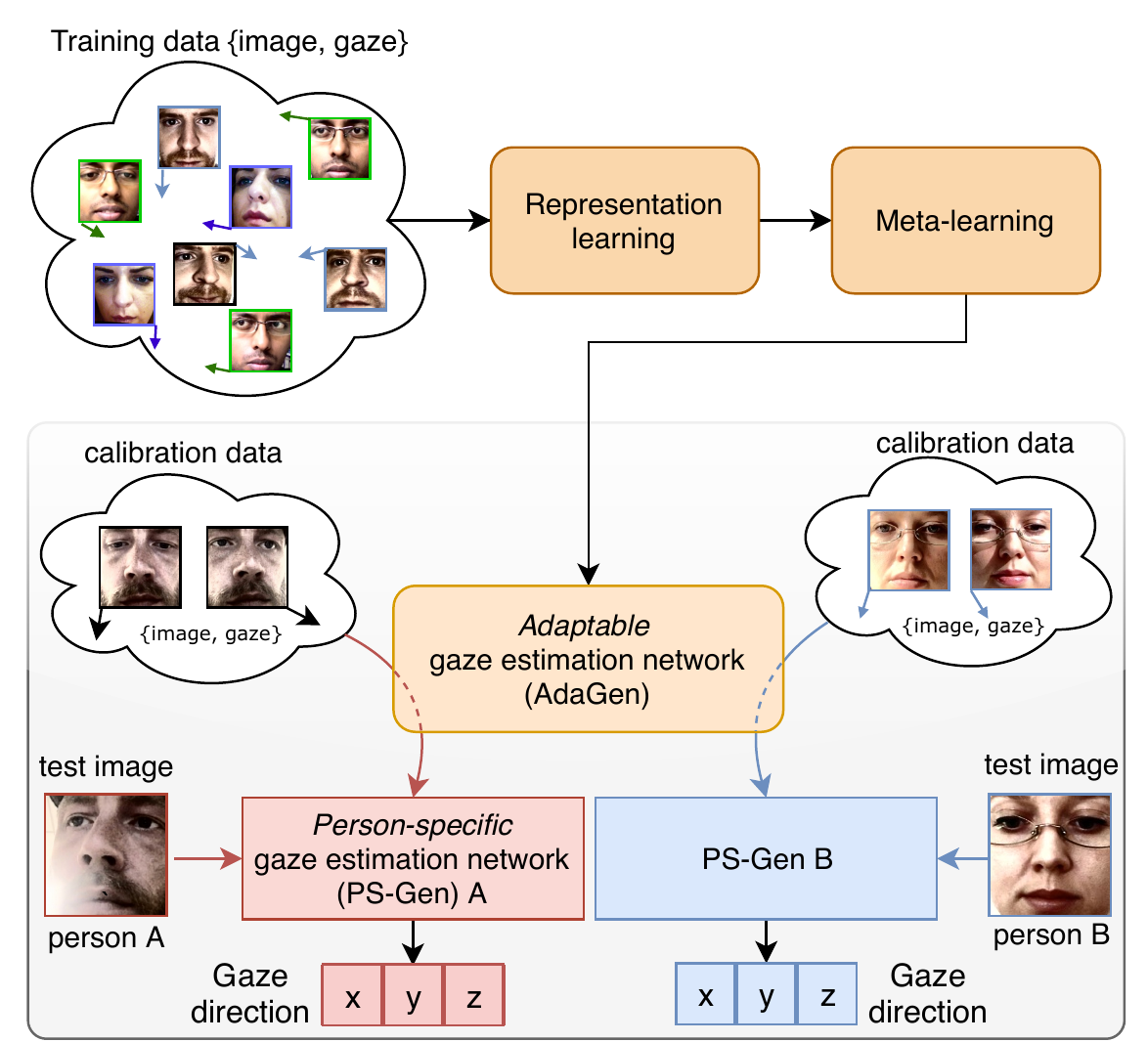}
\captionof{figure}{Overview of the \faze framework. Given a set of training images with ground-truth gaze direction information, we first learn a latent feature representation, which is tailored specifically for the task of gaze estimation. Given the features, we then learn an adaptable gaze estimation network, AdaGEN, using meta-learning which can be adapted easily to a robust person-specific gaze estimation network (PS-GEN) with very little calibration data. 
\label{fig:faze_framework}}
\vspace{-0.2cm}
\end{figure}

Estimation of human gaze has numerous applications in human-computer interaction \cite{Fridman2018CHI}, 
virtual reality \cite{Patney2016SIGGRAPH}, automotive \cite{tawari2014driver} and content creation \cite{Wedel2008}, among others. 
Many of these applications require high levels of accuracy (cf. \cite{Biedert2010CHIEA,Sibert2000UIST,Huang2016MM,Betke2002TNSRE}). 
While appearance-based gaze estimation techniques that use Convolutional Neural Networks (CNN) have significantly surpassed classical methods \cite{Zhang2015CVPR} for in-the-wild settings, there still remains a significant gap towards applicability in high-accuracy domains. The currently lowest reported person-independent  error  of $4.3^\circ$ \cite{Fischer2018ECCV} is equivalent to 4.7cm at a distance of 60cm, which restricts use of such techniques to public display interactions \cite{Zhang2014UbiComp} or crowd-sourced attention analysis \cite{Papoutsaki2017CHIIR}.

High-accuracy gaze estimation from images is difficult because it requires either explicit or implicit fitting of a person-specific eye-ball model to the image data and the estimation of their visual and optical axes. 
Moreover, it is well understood that inter-subject anatomical differences affect gaze estimation accuracy \cite{guestrin2006general}.  
Classical model-based techniques can often be personalized via few (9 or less) samples (\eg \cite{guestrin2006general, Hansen2010TPAMI}) but are not robust to image variations in uncontrolled settings. 
While feasible, subject-specific training of CNNs requires thousands of samples and is clearly impractical \cite{Zhang2019TPAMI}. 
Few-shot personalization of CNNs is difficult because training 
of highly overparametrized models with only few training samples will lead to over-fitting.

We tackle these many-fold challenges by proposing \faze, a framework for learning gaze estimation networks for new subjects using very few calibration samples (Fig.~\ref{fig:faze_framework}).
It consists of:
\begin{inparaenum}[i)]
\item learning a rotation-aware latent representation of gaze via a disentangling transforming encoder-decoder architecture 
\item with these features learning a highly adaptable gaze estimator using meta-learning, and 
\item adapting it to any new person to yield significant performance gains with as few as $3$ samples.
\end{inparaenum}

In order to learn a robust representation for gaze, we take inspiration from recent work on transforming encoder-decoder architectures \cite{Hinton2011ICANN,Worrall2017ICCV} and design a rotation-equivariant pair of encoder-decoder functions. 
We disentangle the factors of appearance, gaze and head pose in the latent space and enforce equivariance by decoding explicitly rotated latent codes to images of the \emph{same} person but with a \emph{different} gaze direction compared to the input (via a $\ell_1$ reconstruction loss).
The equivariance property of our gaze representation further allows us to devise a novel \emph{embedding consistency} loss term that further minimizes the intra-person differences in the gaze representation. 
We then leverage the proposed latent embedding to learn person-specific gaze estimators from few samples. To this end we use a meta-learning algorithm to learn \emph{how to learn} such estimators. We take inspiration from the recent success of meta-learning~\cite{andrychowicz2016learning} for few-shot learning in several other vision tasks \cite{Finn2017ICML, gui2018few, park2018meta}. To the best of our knowledge, we are the first to cast few-shot learning of person-specific gaze estimators as one of  multi-task learning where each subject is seen as a new task in the meta-learning sense.

We evaluate the proposed framework on two benchmark datasets and show that our meta-learned network with its latent gaze features can be successfully adapted with very few ($k\leq9$) samples to produce accurate person-specific models. We demonstrate the validity of our design choices with detailed empirical evidence, and demonstrate that our proposed framework outperforms state-of-the-art methods by significant margins.   
In particular, we demonstrate improvements of $13\%$ ($3.94^\circ\rightarrow 3.42^\circ$) on the MPIIGaze dataset, and $19\%$ ($3.91^\circ\rightarrow 3.18^\circ$) on the GazeCapture dataset over the approach of \cite{Liu2018BMVC} using just $3$ calibration samples.

\noindent To summarize, the main contributions of our work are: 
\begin{compactitem}
\item \faze, a novel framework for learning person-specific gaze networks with few calibration samples, fusing the benefits of rotation-equivariant feature learning and meta-learning.
\item A novel encoder-decoder architecture that disentangles gaze direction, head pose and appearance factors.
\item A novel and effective application of meta-learning to the task of few-shot personalization.
\item State-of-the-art performance ($3.14^{\circ}$ with $k=9$ on MPIIGaze), with consistent improvements over existing methods for $1\leq k\leq 256$. 

\end{compactitem}

\section{Related Work}

\Paragraph{Gaze Estimation.}
Appearance-based gaze estimation \cite{Tan2002WACV} methods that map images directly to gaze have recently surpassed classical model-based approaches \cite{Hansen2010TPAMI} for in-the-wild settings. 
Earlier approaches in this direction assume images captured in restricted laboratory settings and use direct regression methods \cite{Lu2011ICCV,Lu2011BMVC} or learning-by-synthesis approaches combined with random forests to separate head-pose clusters \cite{Sugano2014CVPR}. 
More recently, the availability of large scale datasets such as MPIIGaze \cite{Zhang2015CVPR} and GazeCapture \cite{Krafka2016CVPR}, and progress in CNNs have rapidly moved the field forward. MPIIGaze has become a benchmark dataset for in-the-wild gaze estimation. For the competitive person-independent within-MPIIGaze leave-one-person-out evaluation, gaze errors have progressively decreased from $6.3^\circ$ for naively applying a LeNet-5 architecture to eye-input \cite{Zhang2015CVPR} to the current best of $4.3^\circ$ with an ensemble of multi-modal networks based on VGG-16 \cite{Fischer2018ECCV}.
Proposed advancements include the use of more complex CNNs \cite{Zhang2019TPAMI}; more meaningful use of face \cite{Zhang2017CVPRW, Krafka2016CVPR} and multi-modal input \cite{Krafka2016CVPR,Fischer2018ECCV,Yu2018ECCVW}; explicit handling of differences in the two eyes \cite{Cheng2018ECCV}; greater robustness to head pose \cite{zhu2017monocular, Ranjan2018CVPRW}; improvements in data normalization \cite{Zhang2018ETRA}; learning more informed intermediate representations \cite{Park2018ECCV}; using ensembles of networks \cite{Fischer2018ECCV}; and using synthetic data \cite{Shrivastava2017CVPR, Wang2018CVPR,Lee2018ICLR, Park2018ETRA, Ranjan2018CVPRW}. 

However, person-independent gaze errors are still insufficient for many applications \cite{Biedert2010CHIEA,Sibert2000UIST,Huang2016MM,Betke2002TNSRE}. While significant gains can be obtained by training person-specific models, it requires many thousands of training images per subject \cite{Zhang2019TPAMI}. On the other hand, CNNs are prone to over-fitting  if trained with very few ($k\leq9$) samples. In order to address this issue, existing approaches try to adapt person-independent CNN-based features \cite{Krafka2016CVPR, Park2018ETRA} or points-of-regard (PoR) \cite{Zhang2019CHI} to person-specific ones via hand-designed heuristic functions. Some methods also train a Siamese network with pairs of images of the same subject \cite{Liu2018BMVC}.

\Paragraph{Learned Equivariance.}
Generalizing models learned for regression tasks to new data is a challenging problem.
However, recent works show improvements from enforcing the learning of equivariant mappings between input, latent features, and label spaces \cite{Honari2018CVPR,Rhodin2018ECCV}, via so-called transforming encoder-decoder architectures \cite{Hinton2011ICANN}.
In \cite{Worrall2017ICCV}, this idea is expanded to learn complex phenomena such as the orientation of synthetic light sources 
and in \cite{Rhodin2018ECCV} the method is applied to real-world multi-view imagery to improve semi-supervised human pose estimation.
In contrast, we learn from very noisy real-world data while successfully disentangling the two noisily-labeled phenomena of gaze direction and head orientation.

\Paragraph{Few-shot Learning.} Few-shot learning aims to learn a new task with very few examples~\cite{lake2015human}. This is a non-trivial problem for highly over-parameterized deep networks as it leads to over-fitting. Recently, several promising meta-learning \cite{snell2017prototypical,vinyals2016matching, rezende2016oneshot,santoro2016meta, Finn2017ICML,Nichol2018:reptile,ravi2017metalstm} techniques have been proposed that learn \emph{how to learn} unique but similar tasks in a few-shot manner using CNNs. They have been shown to be successful for various few-shot visual learning tasks including object recognition \cite{Finn2017ICML}, segmentation \cite{rakelly2018few}, viewpoint estimation \cite{Tseng2019BMVC} and online adaptation of trackers \cite{park2018meta}. Inspired by their success, we use meta-learning to learn how to learn person-specific gaze networks from few examples. To the best of our knowledge we are the first to cast person-specific gaze estimation as a multi-task problem in the context of meta-learning, where each subject is seen as a new task for the meta-learner. 
Our insight is that meta-learning lends itself well to few-shot gaze personalization and leads to performance improvements.

\section{Method}

In this section, we describe how we perform gaze estimation from challenging in-the-wild imagery, with minimal burden to the user. The latter objective can be fulfilled by designing our framework to adapt well even with very few calibration samples ($k\leq 9$).
We first provide an overview of the  \faze framework and its three stages.

\subsection{The \faze framework\label{sec:Faze}}
We design \faze (Fig. \ref{fig:faze_framework}) with the understanding that a person-specific gaze estimator must encode factors particular to the person, yet at the same time, leverage insights from observing the eye-region appearance variations across a large number of people with different head pose and gaze direction configurations.
The latter is important for building models that are robust to extraneous factors such as poor image quality.
Thus, the first step in \faze is to learn a generalizable latent embedding space that encodes information pertaining to the gaze-direction, including person-specific aspects.
We detail this in Sec.~\ref{sec:RotAE}.

Provided that good and consistent features can be learned, we can leverage meta-learning to learn how to learn few-shot person-specific gaze estimators for these features.
This results in few-shot learners which generalize better to new people (tasks) without overfitting. 
Specifically, we use the MAML meta-learning algorithm \cite{Finn2017ICML}. 
For our task, MAML learns a set of initial network weights which allow for fine-tuning without the usual over-fitting issues that occur with low $k$. Effectively, it produces a highly Adaptable Gaze Estimation Network (AdaGEN).
The final step concerns the adaptation of MAML-initialized weights to produce person-specific models (PS-GEN) for each user.
We describe this in Sec.~\ref{sec:PS-GEN}.

\subsection{Gaze-Equivariant Feature Learning}\label{sec:RotAE}

In this section, we explain how the learning of a function, which understands equivalent rotations in input data and output label can lead to better generalization in the context of our final task of person-specific gaze estimation.
In addition, we: \begin{inparaenum}[(a)]
\item show how to disentangle eyeball and head rotation factors leading to better distillation of gaze information,
and 
\item introduce a frontalized \emph{embedding consistency} loss term to specifically aid in learning consistent frontal gaze codes for a particular subject.
\end{inparaenum}

\subsubsection{Architecture Overview}
\begin{figure}
    \centering
    \vskip -0.5mm
    \includegraphics[width=0.99\linewidth]{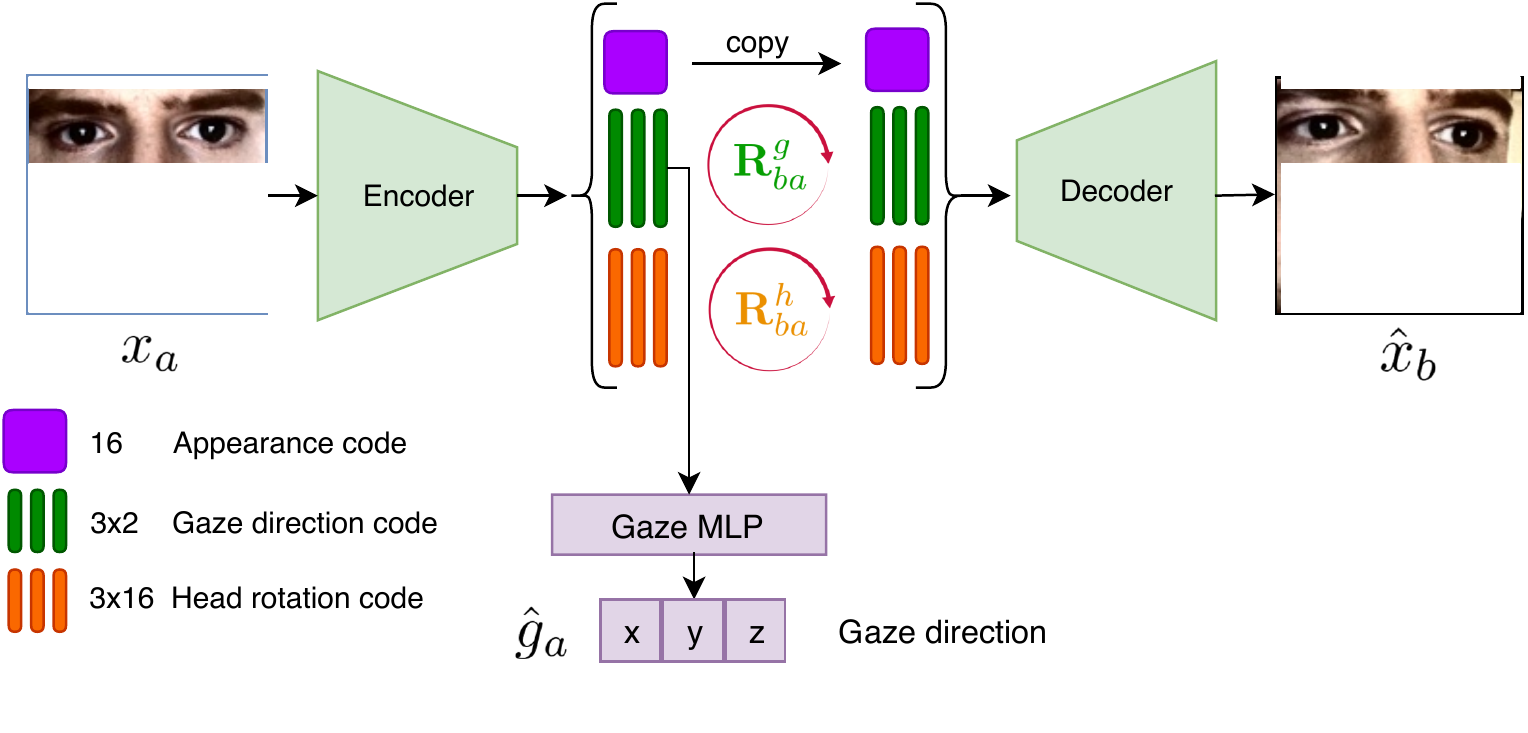}
    \vskip -5.5mm
    \caption{Disentangling appearance, gaze and head pose variations from an image with our Disentangling Transforming Encoder-Decoder (DT-ED). We learn to translate between pairs of images of the same person by rotating the gaze and head pose codes. The encoder-decoder are supervised by a pixel-wise $L_1$ loss (Eq.~\ref{eq:l1_loss}), with the gaze embedding additionally supervised via gaze regression (Eq.~\ref{eq:gaze_loss}).}
    \label{fig:t_ae}
    \vspace*{-2mm}
\end{figure}
In learning a generalizable latent embedding space representing gaze, we apply the understanding that a relative change in gaze direction is easier to learn in a person-independent manner \cite{Liu2018BMVC}.
We extend the transforming encoder-decoder architecture \cite{Hinton2011ICANN,Worrall2017ICCV} to consider three distinct factors apparent in our problem setting: gaze direction, head orientation, and other factors related to the appearance of the eye region in given images (Fig. \ref{fig:t_ae}). 
We disentangle the three factors by \emph{explicitly} applying separate and known differences in rotations (eye gaze and head orientation) to the respective sub-codes.
We refer to this architecture as the Disentangling Transforming Encoder-Decoder (DT-ED).

For a given input image $\mathbf{x}$
, we define an encoder $\mathcal{E}: \mathbf{x}\rightarrow\mathbf{z}$ and a decoder $\mathcal{D}: \mathbf{z}\rightarrow\hat{\mathbf{x}}$ such that $\mathcal{D}\left(\mathcal{E}(\mathbf{x})\right)=\hat{\mathbf{x}}$.
We consider the latent space embedding $\mathbf{z}$ as being formed of 3 parts representing: appearance ($\mathbf{z}^a$), gaze direction or eyeball rotation ($\mathbf{z}^g$), and head pose ($\mathbf{z}^h$), which can be expressed as: $\mathbf{z} = \left\{\mathbf{z}^a;\,\mathbf{z}^g;\,\mathbf{z}^h\right\}$ where gaze and head codes are flattened to yield a single column.
We define $\mathbf{z}^g$ as having dimensions $\left(3\times F^g\right)$ and $\mathbf{z}^h$ as having dimensions $\left(3\times F^h\right)$ with $F \in \mathbb{N}$.
With these dimensions, it is possible to apply a rotation matrix to explicitly rotate these $3D$ latent space embeddings using rotation matrices.

The frontal orientation of eyes and heads in our setting can be represented as 
$\left(0,\,0\right)$ in Euler angles notation for azimuth and elevation, respectively assuming no roll, and using the $x-y$ convention. Then, the rotation of the eyes and the head from the frontal orientation can be described using $\left(\theta^g,\phi^g\right)$ and $\left(\theta^h,\phi^h\right)$ in Euler angles and converted to rotation matrices defined as,
\begin{equation}
    \mathbf{R}^{\left(\theta,\,\phi\right)}
    = \begin{bmatrix}
        \cos{\phi} & 0 & \sin{\phi} \\
        0 & 1 & 0 \\
        -\sin{\phi} & 0 & \cos{\phi} \\
    \end{bmatrix}
    \begin{bmatrix}
        1 & 0 & 0 \\
        0 & \cos{\theta} & -\sin{\theta} \\
        0 & \sin{\theta} & \cos{\theta} \\
    \end{bmatrix}.
    \label{eq:euler2R}
\end{equation}

While training DT-ED, we input a pair of images of a person $\mathbf{x}_a$ and $\mathbf{x}_b$. We can calculate $\mathbf{R}^g_{ba}=\mathbf{R}^g_b\left(\mathbf{R}^g_a\right)^{-1}$ to describe the change in gaze direction in going from sample $a$ to sample $b$ of the same person.
Likewise for head rotation, $\mathbf{R}^h_{ba}=\mathbf{R}^h_b\left(\mathbf{R}^h_a\right)^{-1}$.
This can be done using the \emph{ground-truth} labels for gaze ($\mathbf{g}_a$ and $\mathbf{g}_b$) and head pose ($\mathbf{h}_a$ and $\mathbf{h}_b$) for the pair of input samples.
The rotation of the latent code $\mathbf{z}^g_a$ can then be expressed via the operation $\hat{\mathbf{z}}^g_b=\mathbf{R}_{ab}^g\mathbf{z}^g_a$.
At training time, we enforce this code to be equivalent to the one extracted from image $\mathbf{x}_b$, via a reconstruction loss (Eq.~\ref{eq:l1_loss}). 
We assume the rotated codes $\hat{\mathbf{z}}^h_b$ and $\hat{\mathbf{z}}^g_b$, along with the appearance-code $\mathbf{z}^a_a$, to be sufficient for reconstructing $\mathbf{x}_b$ through the decoder function such that, $\mathcal{D}\left(\hat{\mathbf{z}}_b\right) = \mathbf{x}_b$. 
More specifically, given the encoder output $\mathcal{E}\left(\mathbf{x}_a\right)=\mathbf{z}_a = \left\{\mathbf{z}^a_a;\,\mathbf{z}^g_a;\,\mathbf{z}^h_a\right\}$, we assume the rotated version of $\mathbf{x}_a$ to match the embedding of $\mathbf{x}_b$, that is we assume $\left\{\hat{\mathbf{z}}^a_b;\,\hat{\mathbf{z}}^g_b;\,\hat{\mathbf{z}}^h_b\right\}=\left\{\mathbf{z}^a_a;\,\mathbf{R}^g_{ba}\mathbf{z}^g_a;\,\mathbf{R}^h_{ba}\mathbf{z}^h_a\right\}$ (See Fig.~\ref{fig:t_ae}).

\begin{figure}[t]
    \centering
    \begin{subfigure}[t]{\columnwidth}
        \includegraphics[width=1.0\textwidth]{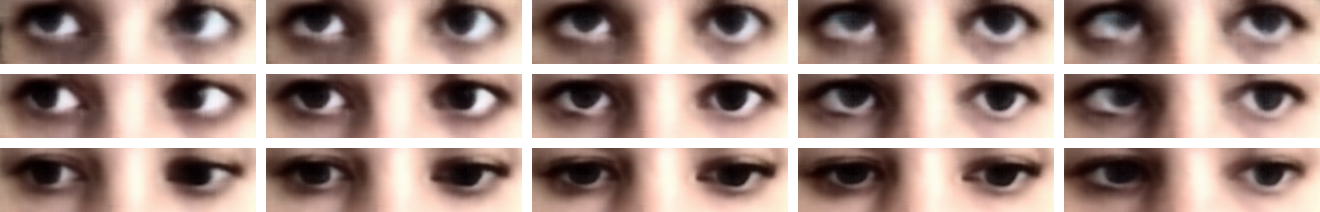} 
        \vskip -1.8mm
        \caption{\small Only varying gaze direction, $\left(\theta^g,\,\phi^g\right)\in[-25^\circ,\,25^\circ]$}
    \end{subfigure}
    \begin{subfigure}[t]{\columnwidth}
        \includegraphics[width=1.0\textwidth]{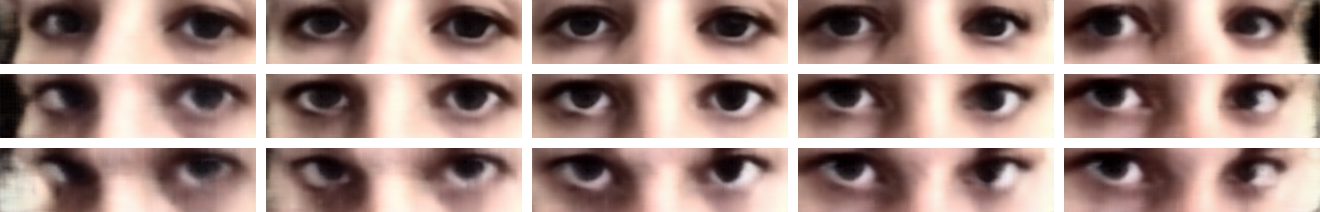}
        \vskip -1.8mm
        \caption{\small Only varying head orientation, $\left(\theta^h,\,\phi^h\right)\in[-20^\circ,\,20^\circ]$}
    \end{subfigure}
    \vskip -2.2mm
    \caption{Our disentangled rotation-aware embedding spaces for gaze direction and head pose are demonstrated by predicting embeddings $\hat{\mathbf{z}}^g$, $\hat{\mathbf{z}}^h$ from a given sample, rotating it to 15 points each, and then decoding them. 
    \label{fig:latent_walks}
    }
    \vspace*{-5mm}
\end{figure}

This approach indeed applies successfully to noisy real-world imagery, as shown in Fig.~\ref{fig:latent_walks} where we map a sample into the gaze and head pose latent spaces, rotate to the frontal orientation, and then again rotate by a pre-defined set of 15 yaw and pitch values and reconstruct the image via the decoder. We can see that the factors of gaze direction and head pose are fully disentangled
and DT-ED succeeds in the challenging task of eye-region frontalization and re-direction from monocular RGB input.

We train the \faze transforming encoder-decoder architecture using a multi-objective loss function defined as,
\begin{equation}
\label{eq:loss}
    \mathcal{L}_\mathrm{full} = 
    \lambda_\mathrm{recon} \mathcal{L}_\mathrm{recon}
    + \lambda_\mathrm{EC} \mathcal{L}_\mathrm{EC}
    + \lambda_\mathrm{gaze} \mathcal{L}_\mathrm{gaze},
\end{equation}
where we empirically set $\lambda_\mathrm{recon}=1$, $\lambda_\mathrm{EC}=2$, and $\lambda_\mathrm{gaze}=0.1$.
The individual loss terms are explained in the following sub-sections.

\subsubsection{Reconstruction Loss}
To guide learning of the encoding-decoding process, we apply a simple $\ell_1$ reconstruction loss.
Given an input image $\mathbf{x}_b$ and reconstructed $\hat{\mathbf{x}}_b$ obtained by decoding the rotated embeddings $\hat{\mathbf{z}}_b$ of image $\mathbf{x}_a$, the loss term is defined as,
\begin{equation}
    \mathcal{L}_\mathrm{recon}\left(
    \mathbf{x}_b,\,
    \hat{\mathbf{x}}_b
    \right)
    =
    \frac{1}{|\mathbf{x}_b|}
    \sum_{u\in\mathbf{x}_b,\hat{u}\in\hat{\mathbf{x}}_b}
    \left|
        \hat{u} - u
    \right|,
    \label{eq:l1_loss}
\end{equation}
where $u$ and $\hat{u}$ are pixels of images $\mathbf{x}_b$ and $\hat{\mathbf{x}}_b$ respectively.

\subsubsection{Embedding consistency Loss}
\label{sec:embedding consistency}

We introduce a novel embedding consistency term, which ensures that the encoder network always embeds images of a person with different appearance but identical gaze direction to similar features. 
Usually this would require paired images with only gaze directions changed. However, it is intractable to collect such data in the real world, so we instead exploit
the learned equivariance of DT-ED.
Before measuring the consistency between latent gaze features from different samples, we first frontalize them by applying the inverse of the rotation matrix $\mathbf{R}^g_a$ using ground-truth gaze direction $\mathbf{g}_a$. It should be noted that naively enforcing all gaze features to be similar across persons may disregard the inter-subject anatomical differences which should result in different latent embeddings. Hence, we apply the embedding consistency to intra-subject pairs of images only. 
We validate this choice through experiments in Sec.~\ref{sec:ablation}.

Given a batch of $B$ image samples during training, we formally compute the \emph{embedding consistency} loss using,
\begin{equation}
    \mathcal{L}_\mathrm{EC} = 
    \frac{1}{B}
    \sum_{i=1}^{B}
    \max_{\substack{j=1\ldots B\\id(i) = id(j)}}
        d\left( f(\mathbf{z}^g_i),f(\mathbf{z}^g_j) \right),
    \label{eq:embedding consistency_loss}
\end{equation}
where $f(\mathbf{z}^g) = \left(\mathbf{R}^g\right)^{-1}\mathbf{z}^g$ corresponds to frontalized latent gaze features. The function $id(i)$ provides the person-identity of the $i$-th sample in the batch, and $d$ is a function based on mean column-wise angular distance (between 3D vectors).
The $\max$ function minimizes differences between intra-person features that are furthest apart, and is similar to the idea of batch-hard online triplet mining \cite{Schroff2015CVPR}.

During training, we linearly increase $\lambda_\mathrm{EC}$ from $0$ until sufficient mini-batches to cover $10^6$ images have been processed, to allow for more natural embeddings to arise before applying consistency.

\subsubsection{Gaze Direction Loss}
Lastly, we add the additional objective of gaze estimation via $\mathcal{G}: \mathbf{z}^g \rightarrow \hat{\mathbf{g}}$, parameterized by a simple multi-layer perceptron (MLP).
The gaze direction loss is calculated using,
\begin{equation}
    \mathcal{L}_\mathrm{gaze}\left(\hat{\mathbf{g}},\,\mathbf{g}\right)
    =\arccos{\left(\frac{\hat{\mathbf{g}}\cdot\mathbf{g}}{\|\hat{\mathbf{g}}\|\|\mathbf{g}\|}\right)}.
    \label{eq:gaze_loss}
\end{equation}

\subsection{Person-specific Gaze Estimation \label{sec:PS-GEN}}
Having learned a robust feature extractor, which is tailored specifically for gaze estimation, our final goal is to learn a personalized gaze estimator 
with as few calibration samples as possible. A straightforward solution for doing this is to train a person-independent model with the training data used to train DT-ED and simply fine-tune it with the available calibration samples for the given subject. However, in practical setups where only a few calibration samples are available, this approach can quickly lead to over-fitting 
(see experiments in Fig.~\ref{fig:maml-vs-nomaml}). In order to alleviate this problem, we propose to use the meta-learning method MAML~\cite{Finn2017ICML}, which learns a highly adaptable gaze network (AdaGEN).

\paragraph{Adaptable Gaze Estimator (AdaGEN) Training.\label{sec:MAML}}
We wish to learn weights $\theta^*$ for the AdaGEN gaze prediction model $\mathcal{M}$ such that it becomes a  successful few-shot learner. In other words, when $\mathcal{M}_{\theta^*}$ is fine-tuned with only a few ``calibration" examples of a new person $\mathcal{P}$ not present in the training set, it can generalize well to unseen ``validation" examples of the same person. We achieve this by training it with the MAML  meta learning algorithm adapted for few-shot learning. 

In conventional CNN training the objective is to minimize the training loss for all the examples of all training subjects. In contrast, for few-shot learning, MAML explicitly minimizes the \emph{generalization} loss of a network \emph{after} minimizing its training loss for a few examples of a particular subject via a standard optimization algorithm, \eg stochastic gradient descent (SGD). Additionally, MAML repeats this procedure for all subjects in the training set and hence learns from several closely related ``tasks" (subjects) to become a successful few shot learner for any new unseen task (subject). We identify that person-specific factors may have few parameters, with only slight but important variations across people such that all people constitute a set of closely related tasks. Our insight is that meta-learning lends itself well to such a problem of personalization.

The overall procedure of meta-learning to learn the optimal $\theta^*$ is as follows. 
We divide the entire set of persons~$\mathcal{S}$ into meta-training ($\mathcal{S}^{train}$) and meta-testing ($\mathcal{S}^{test}$) subsets of non-overlapping subjects.
During each meta-training iteration $n$, we randomly select one person $\mathcal{P}^{train}$ from $\mathcal{S}^{train}$ and create a meta-training sample for the person (via random sampling), defined as $\mathcal{P}^{train}=\{\mathcal{D}_c^{train},\mathcal{D}_v^{train}\}$, containing a calibration set $\mathcal{D}_c^{train} = \{(\mathbf{z^g}_i, \mathbf{g}_i) | i = 1 \ldots k\}$ of $k$ training examples, and a validation set $\mathcal{D}_v^{train} = \{(\mathbf{z^g}_j, \mathbf{g}_j) | j = 1 \ldots l\}$ of another $l$ examples for the same person.
Here, $\mathbf{z^g}$ and $\mathbf{g}$ refer to the latent gaze representation learned by DT-ED and the ground-truth 3D gaze vector, respectively. 
Both $k$ and $l$ are typically small ($\leq20$) and $k$ represents the ``shot" size used in few-shot learning.

The first step in the meta-learning procedure is to compute the loss for the few-shot calibration set $\mathcal{D}_c^{train}$ and update the weights $\theta_n$ at step $n$ via one (or more) gradient steps and a learning rate $\alpha$ as,
\begin{equation} \label{eq:update1}
    \theta'_n = f(\theta_n) = \theta_n - \alpha \nabla\mathcal{L}_{\mathcal{P}^{train}}^c(\theta_n).
\end{equation}
With the updated weights $\theta'_n$, we then compute the loss for the validation set $\mathcal{D}_v^{train}$ of the subject $\mathcal{P}^{train}$  as $\mathcal{L}_{\mathcal{P}^{train}}^v(\theta'_n) = \mathcal{L}_{\mathcal{P}^{train}}^v(f(\theta_n))$ and its gradients w.r.t the initial weights of the network $\theta_n$ at that training iteration $n$. Lastly, we update $\theta_n$ with a learning rate of $\eta$ to explicitly minimize the validation loss as,
\begin{equation} \label{eq:update2}
    \theta_{n+1} = \theta_n - \eta \nabla\mathcal{L}_{\mathcal{P}^{train}}^{v}(f(\theta_n)).
\end{equation}

We repeat these training iterations until convergence to get the optimal weights $\theta^*$.

\paragraph{Final Person-specific Adaptation.}
Having learned our encoder and our optimal few-shot person-specific learner $\mathcal{M}_{\theta^*}$, we are now well poised to produce person-specific models for each new person $\mathcal{P}^{test}$ from $\mathcal{S}^{test}$. We fine-tune $\mathcal{M}_{\theta^*}$ with the $k$ calibration images $\mathcal{D}_c^{test}$ to create a personalized model for $\mathcal{P}^{test}$ as
\begin{equation} \label{eq:update_final}
    \theta_{\mathcal{P}^{test}} = \theta^* - \alpha \nabla\mathcal{L}_{\mathcal{P}^{test}}^{c}(\theta^*), 
\end{equation}
and test the performance of the personalized model $\mathcal{M}_{\theta_{\mathcal{P}^{test}}}$ on person $\mathcal{P}^{test}$'s validation set $\mathcal{D}_v^{test}$.

 \section{Implementation Details}

\subsection{Data pre-processing\label{sec:pre-processing}}
Our data pre-processing pipeline is based on \cite{Zhang2018ETRA}, a revision of the data normalization procedure introduced in \cite{Sugano2014CVPR}. In a nutshell, the data normalization procedure ensures that a common virtual camera points at the same reference point in space with the head upright. This requires the rotation, tilt, and forward translation of the virtual camera. Please refer to \cite{Zhang2018ETRA} for a formal and complete description, and our supplementary materials for a detailed list of changes.

\subsection{Neural Network Configurations}
\Paragraph{DT-ED.}
The functions $\mathcal{E}$ and $\mathcal{D}$ in our transforming encoder-decoder architecture can be implemented with any CNN architecture. We select the DenseNet architecture \cite{Huang2017CVPR} both for  our DT-ED as well as for our re-implementation of state-of-the-art person-specific gaze estimation methods \cite{Liu2018BMVC,Zhang2019CHI}.
The latent codes $\mathbf{z}_a$, $\mathbf{z}_g$, and $\mathbf{z}_h$ are defined to have dimensions $(64)$, $(3\times 2)$, and $(3\times 16)$ respectively.
Please refer to supplementary materials for further details.

\Paragraph{Gaze MLP.}
Our gaze estimation function $\mathcal{G}$ is parameterized by a multi-layer perceptron with $64$ hidden layer neurons and SELU \cite{Klambauer2017NeurIPS} activation.
The MLP outputs 3-dimensional unit gaze direction vectors.

\subsection{Training}
\Paragraph{DT-ED.}
Following \cite{Goyal2017arXiv}, we use a batch size of $1536$ and apply linear learning rate scaling and ramp-up for the first $10^6$ training samples.
We use NVIDIA's Apex library\footnote{\url{https://github.com/NVIDIA/apex}} for mixed-precision training. 
and train our model for $50$ epochs with a base learning rate of $5\times 10^{-5}$, $l_2$ weight regularization  of $10^{-4}$, and use instance normalization \cite{Ulyanov2016arXiv}. 

\Paragraph{Gaze MLP.}
During meta-learning, we use $\alpha = 10^{-5}$ with SGD (Eq. \ref{eq:update1}), and  $\eta = 10^{-3}$ (Eq. \ref{eq:update2}) with the Adam optimizer ($\alpha$ and $\beta$ in \cite{Finn2017ICML}), and do $5$ updates per inner loop iteration. For sampling $\mathcal{D}^{train}_v$ we set $l=100$. 
During standard eye-tracker calibration, one cannot assume the knowledge of extra ground-truth beyond the $k$ samples.
Thus, we perform the final fine-tuning operation (Eq. \ref{eq:update_final}) for $1000$ steps for all values of $k$ and for all people.

\subsection{Datasets}

\Paragraph{GazeCapture \cite{Krafka2016CVPR}} is the largest available in-the-wild gaze dataset. We mined camera intrinsic parameters from the web for the devices used, and applied our pre-processing pipeline (Sec.~\ref{sec:pre-processing}) to yield input images.
For training the DT-ED as well as for meta-learning, we use data from $993$ people in the training set specified in \cite{Krafka2016CVPR}, each with $1766$ samples, on average, for a total of $1.7M$ samples. 
To ensure within-subject diversity of sampled image-pairs at training time, we only select subjects with $\geq 400$ samples.
For computing our final evaluation metric, we use the last $500$ entries from $109$ subjects that have at least $1000$ samples each. We select the $k$-shot samples for meta-training and fine-tuning randomly from the remaining samples.

\Paragraph{MPIIGaze \cite{Zhang2015CVPR}} is the most established benchmark dataset for in-the-wild gaze estimation. In comparison to GazeCapture it has higher within-person variations in appearance including illumination, make-up, and facial hair changes, potentially making it more challenging.
We use the images specified in the MPIIFaceGaze subset \cite{Zhang2017CVPRW} only for evaluation purposes. The MPIIFaceGaze subset consists of $15$ subjects each with $2500$ samples on average. We reserve the last $500$ images of each subject for final evaluations as is done in \cite{Zhang2019TPAMI} and select $k$ calibration samples for personalization by sampling randomly from the remaining samples. 
 
\section{Results}

For all methods, we report person-specific gaze estimation errors for a range of $k$ calibration samples.
For each data point, we perform the evaluation $10$ times using $k$ randomly chosen calibration samples. 
Each evaluation or trial yields a mean gaze estimation error in degrees over all subjects in the test set.
The mean error over all such trials is plotted, with its standard deviation represented by the shaded areas above and below the curves.
The values at $k = 0$ are determined via $\mathcal{G}\left(\mathbf{z}^g\right)$. 
We train this MLP on the GazeCapture training subset, without any person-specific adaptation.

\subsection{Ablation Study}
\label{sec:ablation}

\begin{figure*}
    \centering
    \begin{subfigure}[b]{0.33\textwidth}
        \includegraphics[width=\columnwidth]{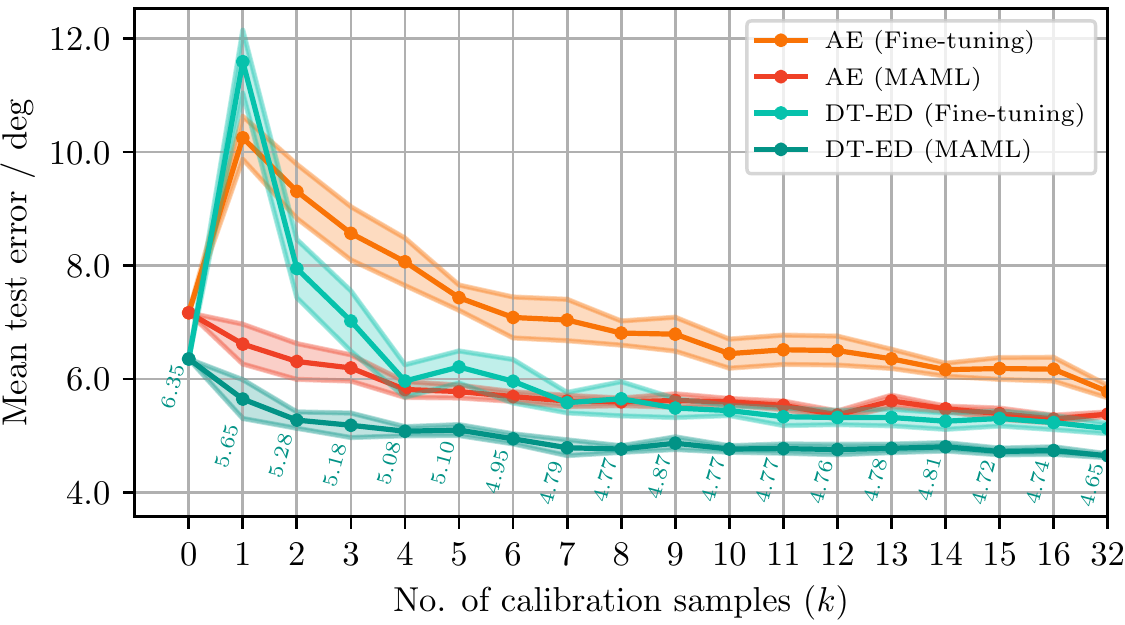}
        \vskip -1mm
        \caption{\vspace{-3mm}}
        \label{fig:maml-vs-nomaml}
    \end{subfigure}
    \hfill
    \begin{subfigure}[b]{0.33\textwidth}
        \includegraphics[width=\columnwidth]{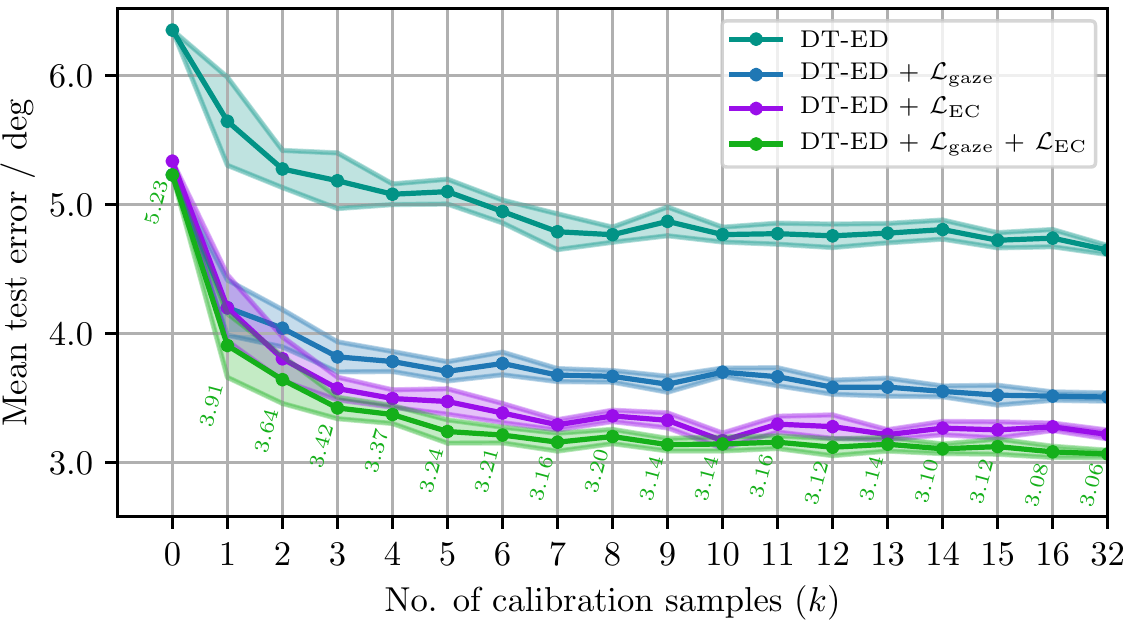}
        \vskip -1mm
        \caption{\vspace{-3mm}}
        \label{fig:impact_of_loss_terms}
    \end{subfigure}
    \begin{subfigure}[b]{0.33\textwidth}
        \includegraphics[width=\columnwidth]{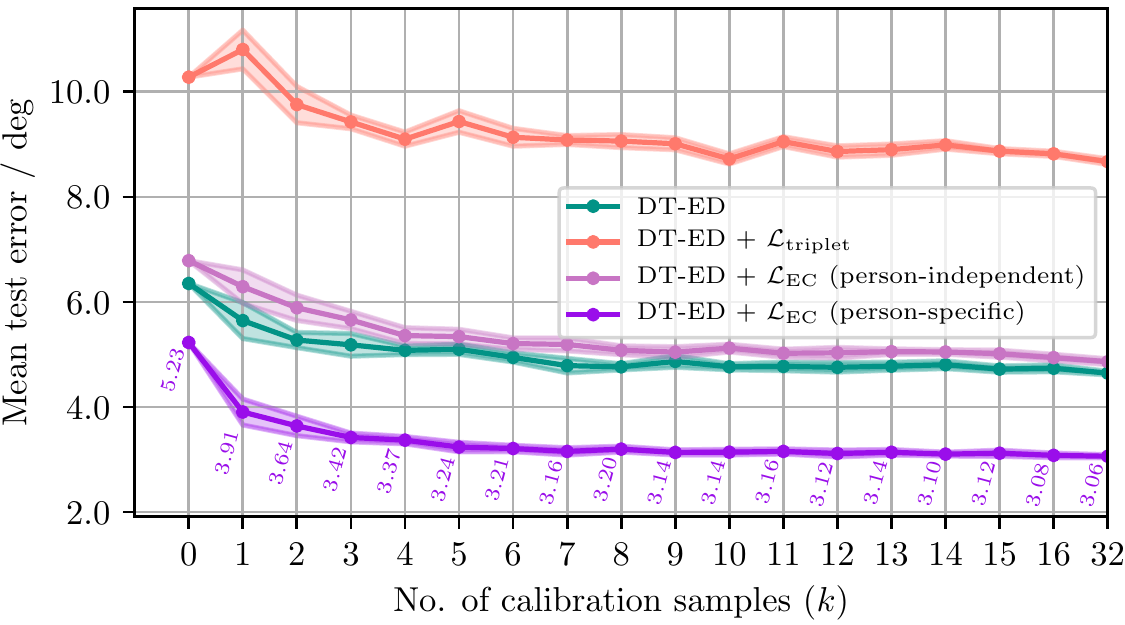}
        \vskip -1mm
        \caption{\vspace{-3mm}}
        \label{fig:impact_of_embedding consistency}
    \end{subfigure}
    \caption{\textbf{Ablation Study:} Impact of (a) learning the few-shot gaze estimator using MAML (Sec.~\ref{sec:MAML}) and using the transforming encoder-decoder for feature learning (Sec.~\ref{sec:RotAE}); (b) different loss terms in Eq. \eqref{eq:loss} for training the transforming encoder-decoder; and (c) comparison of the different variants of embedding consistency loss term (Eq. \eqref{eq:embedding consistency_loss}). 
    We provide additional results for the test partition of the GazeCapture dataset in the supplementary material. 
    \vspace{-5mm}
    }
    \label{fig:ablation}
\end{figure*}

We evaluate our method under different settings to better understand the impact of our various design choices. For all experiments, we train the models using the GazeCapture dataset's training set and test on the MPIIGaze dataset. This challenging experiment allows us to demonstrate the generalization capability of our approach across different datasets. The ablation studies are summarized in Fig.~\ref{fig:ablation}. We provide additional plots of the results of this ablation study on the test partition of the GazeCapture dataset in the supplementary material.

\Paragraph{MAML~vs.~Finetuning.} In  Fig.~\ref{fig:maml-vs-nomaml}, we first evaluate the impact of meta-learning a few-shot person-adaptive gaze estimator using MAML (Sec.~\ref{sec:MAML}) and compare its performance with naive finetuning. When no person-specific adaptation is performed (\ie $k=0$), the person-independent baseline model $\mathcal{G}\left(\mathbf{z}^g\right)$ with the features learned using a vanilla autoencoder (AE) 
results in a mean test error of $7.17\degree$. Using MAML for person-specific adaptation with only one calibration sample decreases the error to $6.61\degree$. The error reduces further as we increase $k$ and reaches a mean value of $5.38\degree$ for $k=32$. 
In contrast, naively finetuning (AE-Finetuning) the gaze estimator results in severe over-fitting and yields very high test errors, in particular, for very low $k$ values. In fact, for $k \le 3$, the error increases to above the person-independent baseline model. Since the model initialized with weights learned by MAML clearly outperforms vanilla finetuning, in the rest of this section, we always use MAML unless specified otherwise.

\Paragraph{Impact of feature representation.} Fig.~\ref{fig:maml-vs-nomaml} also evaluates the impact of the features used to learn the gaze estimation model. Our proposed latent gaze features (Sec.~\ref{sec:RotAE}) significantly decrease the error, \eg $4.87\degree$ vs. $5.62\degree$ with $k=9$ for DT-ED (MAML) and AE (MAML), respectively. Note that the gain remains consistent across all values of $k$. 
The only difference between DT-ED and AE is that the latent codes are rotated in DT-ED before decoding.
Despite this more difficult task, the learned code clearly better informs the final task of person-specific gaze estimation, showing that disentangling gaze, head pose, and appearance is importance for gaze estimation.

\Paragraph{Contribution of loss terms.} We evaluate the impact of each loss term described in Eq. \eqref{eq:loss} (Sec.~\ref{sec:RotAE}) by incorporating them one at a time into the total loss used to train DT-ED. Fig.~\ref{fig:impact_of_loss_terms} summarizes the results. Using only the image reconstruction loss $\mathcal{L_{\mathrm{recon}}}$ in Eq. \eqref{eq:l1_loss}, the learned latent gaze features result in an error of $4.87\degree$ at $k=9$. Incorporating gaze supervision $\mathcal{L}_{\mathrm{gaze}}$ in Eq. \eqref{eq:gaze_loss} to obtain features that are more informed of the ultimate task of gaze-estimation gives an improvement of $26\%$ from $4.87\degree$ to $3.60\degree$. Adding the person-specific embedding consistency term  $\mathcal{L}_{\mathrm{EC}}$ in Eq. \eqref{eq:embedding consistency_loss} to $\mathcal{L}_{\mathrm{recon}}$ also reduces the error significantly from $4.87\degree$ to $3.32\degree$ at $k=9$ (an improvement of over $30\%$). 
Finally, combining all loss terms improves the performance even further
to $3.14\degree$ (in total, an improvement of $36\%$).

\Paragraph{Analysis of embedding consistency.} In order to validate our choice of the embedding consistency loss, in Fig.~\ref{fig:impact_of_embedding consistency}, we compare its performance with two other possible variants. As described in Sec.~\ref{sec:embedding consistency}, the embedding consistency loss term minimizes the intra-person differences of the frontalized latent gaze features. The main rationale behind this is that the gaze features for a unique person should be consistent while they can be different across subjects due to inter-subject anatomical differences. We further conjecture that preserving these inter-personal differences as opposed to trying to remove them by learning \emph{person-invariant} embeddings is indeed important to obtaining high accuracy for gaze estimation. In order to validate this observation, we introduce a person-\emph{independent} embedding consistency term which also minimizes the inter-person latent gaze feature differences. As is evident from Fig.~\ref{fig:impact_of_embedding consistency}, enforcing person-independent embedding consistency of the latent gaze features results in increased mean errors. In fact it performs worse than only using the reconstruction loss ($\mathcal{L}_{\mathrm{recon}}$). 

\begin{figure}
    \centering
    \includegraphics[width=1.0\columnwidth]{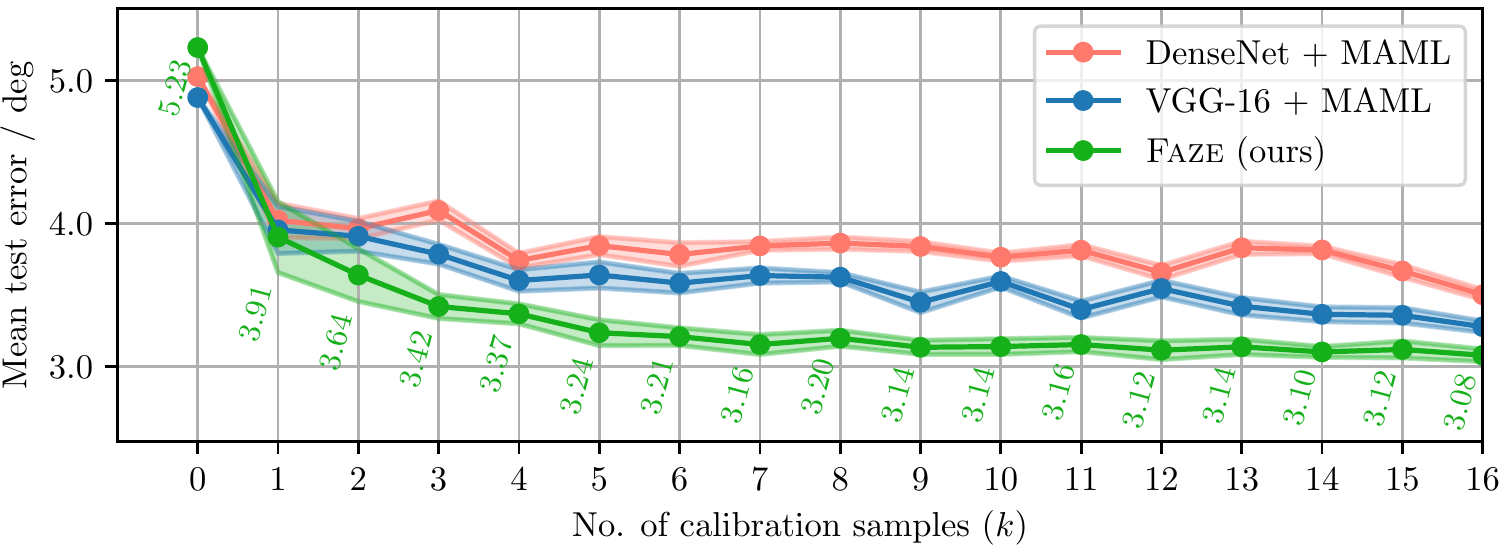}
    \vskip -1mm
    \caption{Comparison of \faze against competitive CNN + MAML baselines, evaluated on MPIIGaze.
    }
    \label{fig:cnn-maml}
    \vskip -3mm
\end{figure}

One may argue the complete opposite \ie the latent gaze features should be hugely different for every subject for the best possible subject-specific accuracy, but we did not find this to be the case. To demonstrate this, we apply a triplet loss ($\mathcal{L}_{\mathrm{triplet}}$)~\cite{Schroff2015CVPR}, which explicitly \emph{maximizes} the inter-personal differences in gaze features in addition to minimizing the intra-person ones. As is evident from Fig~\ref{fig:impact_of_embedding consistency} this results in a significant increase in the error. This suggests that perhaps factors that quantify the overall appearance of a person's face and define their unique identity may not necessarily be correlated to the anatomical properties that define ``person-uniqueness" for the task of gaze estimation. 

\begin{figure*}
    \centering
    \begin{subfigure}[b]{0.48\textwidth}
        \includegraphics[width=\textwidth]{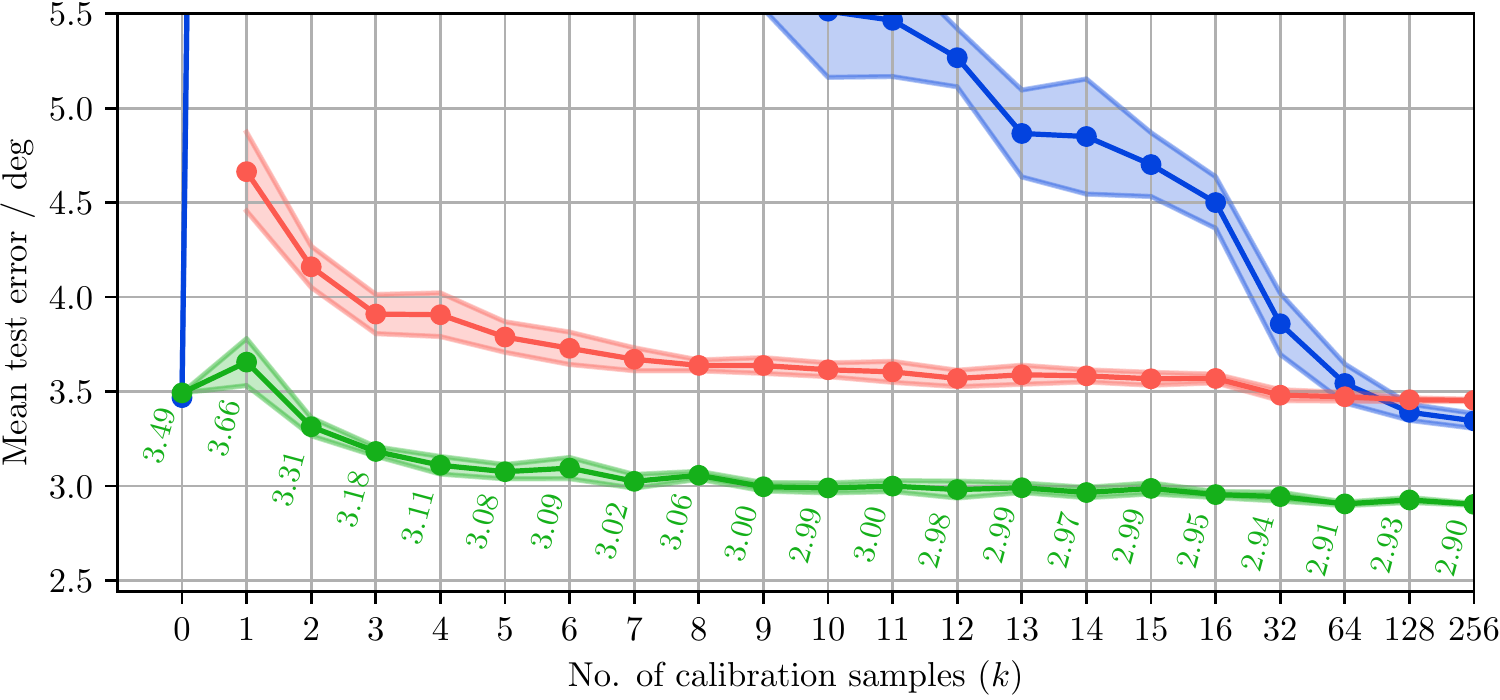}
        \vskip -1mm
        \caption{GazeCapture (test)}
        \label{fig:sota_gc}
    \end{subfigure}
    \hspace*{\fill}
    \begin{subfigure}[b]{0.48\textwidth}
        \includegraphics[width=\textwidth]{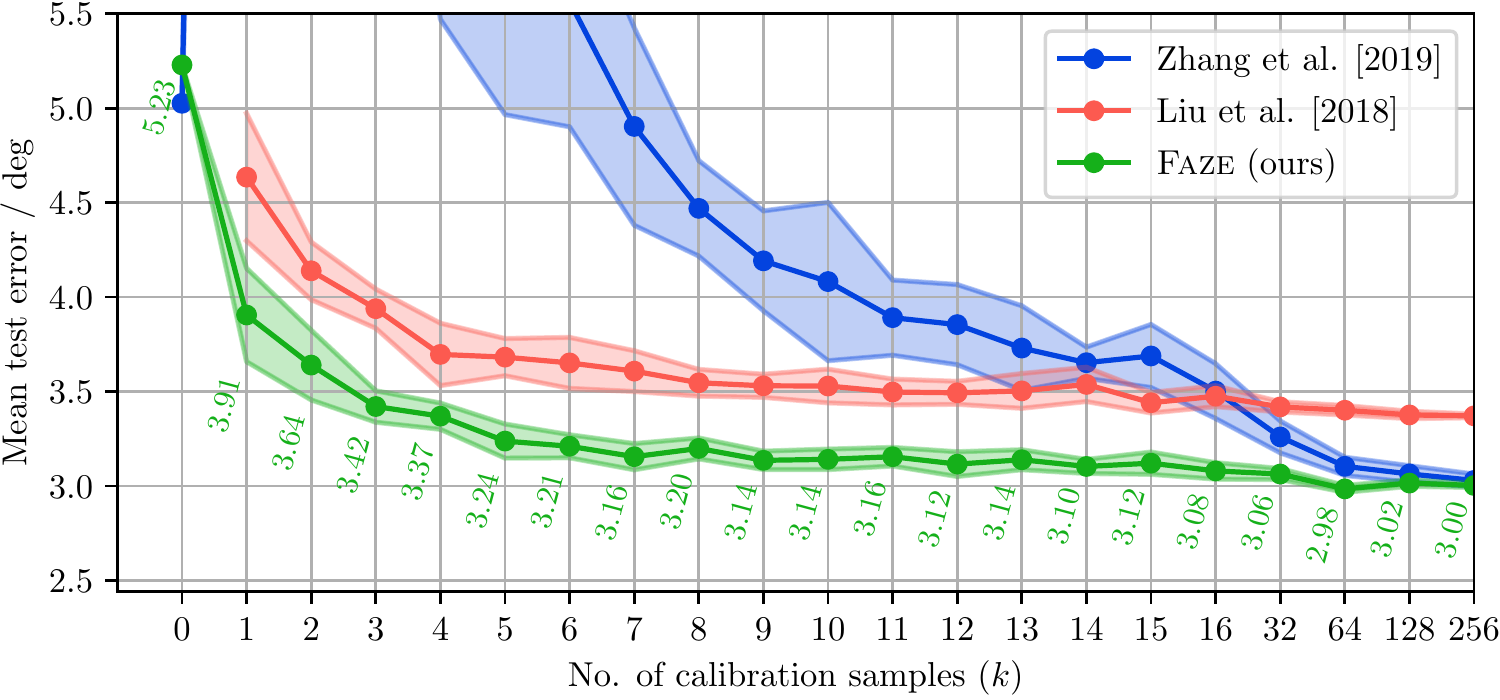}
        \vskip -1mm
        \caption{MPIIGaze}
        \label{fig:sota_mpi}
    \end{subfigure}
    \vskip -2mm
    \caption{Comparison of \faze against state-of-the-art person-specific gaze estimation methods \cite{Liu2018BMVC,Zhang2019CHI}
    }
    \label{fig:sota}
    \vskip -3mm
\end{figure*}

\subsection{Comparison with CNN + Meta-Learning}
An alternative baseline to \faze can be created by replacing the DT-ED with a standard CNN 
architecture.
We take an identically configured DenseNet (to \faze) and a VGG-16 architecture for the convolutional layers, then add $2$ fully-connected layers each with $256$ neurons and train the networks with the gaze objective (Eq.~\ref{eq:gaze_loss}).
The output of the convolutional layers are used as input to a gaze estimation network trained via MAML to yield the results in Fig.~\ref{fig:cnn-maml}.
Having been directly trained on the (cross-person) gaze estimation objective, it is expected that the encoder network would make better use of its model capacity as it does not have to satisfy a reconstruction objective. Thus, we can call these highly competitive baselines.
\faze outperforms these baselines with statistical significance, demonstrating that the DT-ED training and our loss terms yield features which are more amenable to meta-learning, and thus to the final objective of personalized gaze estimation.

\subsection{Comparison with State-of-the-Art}

Few-shot personalization of CNN models in the context of gaze estimation for very low $k$ is very challenging.
Two recent approaches \cite{Zhang2019CHI,Liu2018BMVC} are the most relevant in this direction, and we provide evaluations on highly competitive re-implementations.
Our results are presented in Fig.~\ref{fig:sota} for both the test partition of the GazeCapture dataset and the MPIIGaze dataset.
Overall, we show statistically significantly better mean errors over the entire range of $1\leq k\leq 256$ than all the existing state-of-the-art methods.
In addition, our performance between trials is more consistent as shown by the narrower error bands. This indicates robustness to the choice of the $k$ calibration samples.

\Paragraph{Ours vs Polynomial fit to PoR.} In \cite{Zhang2019CHI}, Zhang et al. fit a 3rd order polynomial function to correct initial point-of-regard (PoR) 
estimates from a person-independent gaze CNN.
To re-implement their method, we train a DenseNet CNN (identical to \faze) and intersect the predicted gaze ray (defined by gaze origin and direction in 3D with respect to the original camera) with the $z=0$ plane to estimate the initial PoR and later refine it with a person-specific 3rd order polynomial function.
Though this approach performs respectably with $k=9$, yielding $4.19^\circ$ on MPIIGaze (Fig.~\ref{fig:sota_mpi}), it suffers with lower $k$ especially on GazeCapture.
Nonetheless, its performance converges to our performance at $k\geq 128$ showing its effectiveness at higher $k$ despite its apparent simplicity.

\Paragraph{Ours vs Differential Gaze Estimation.}
Liu et al. \cite{Liu2018BMVC} introduce a CNN architecture for learning to estimate the differences in the gaze yaw and pitch values between pairs of images of the same subject.
That is, in order to estimate the gaze their network always requires one \emph{reference} image of a subject with known gaze values. Then given a reference image $I_a$ with a known gaze label $\mathbf{g}_a$ and another image $I_b$ with unknown gaze label, their approach outputs a $\Delta\mathbf{g}_{ba}$, from which the absolute gaze for $I_b$ can be computed as $\hat{\mathbf{y}}_b=\mathbf{y}_a+\Delta\mathbf{g}_{ba}$.
Their original paper states a within-MPIIGaze error with $k=9$ at $4.67^\circ$ using a simple LeNet-5 style Siamese network and a pair of eye images as input.
We use $256\times 64$ eye-region images from GazeCapture as input and use a DenseNet-based architecture to make their approach more comparable to ours.
Our re-implementation yields $3.53^\circ$ for their method at $k = 9$ on MPIIGaze, a $1.2^\circ$ improvement despite dataset differences.
We show statistically significant improvements to \cite{Liu2018BMVC} across all ranges of $k$ in our MPIIGaze evaluations, with our method only requiring $4$ calibration samples to compete with their best performance at $k=256$ (see the red and green curves in Fig. \ref{fig:sota}).
The improvement from our final approach is further emphasized in Fig.~\ref{fig:sota_gc} with evaluations on the test subset of GazeCapture.
At $k=4$, we yield a performance improvement of $20.5\%$ or $0.8^\circ$ over \cite{Liu2018BMVC}.
 
\section{Conclusion}
In this paper we presented the first practical approach to deep-learning based high-accuracy personalized gaze estimation requiring only few calibration samples. Our \faze framework consists of a disentangling encode-decoder network that learns a compact person-specific latent representation of gaze, head pose and appearance. Furthermore, we show that these latent embeddings can be used in a meta-learning context to learn a person-specific gaze estimation network from very few (as low as $k=3$) calibration points. We experimentally showed that our approach outperforms other state-of-the-art approaches by significant margins and produces the currently lowest reported personalized gaze errors on both the GazeCapture and MPIIGaze datasets.

\paragraph{Acknowledgements.}
Seonwook Park carried out this work during his internship at Nvidia.
This work was supported in part by the ERC Grant OPTINT (StG-2016-717054). 
\clearpage
\balance
{\small
\bibliographystyle{ieee_fullname}
\bibliography{egbib}
}

\clearpage
\balance
\begin{figure*}[]
    \vspace*{3mm}
    \centering
    \begin{subfigure}[b]{0.33\textwidth}
        \includegraphics[width=\columnwidth]{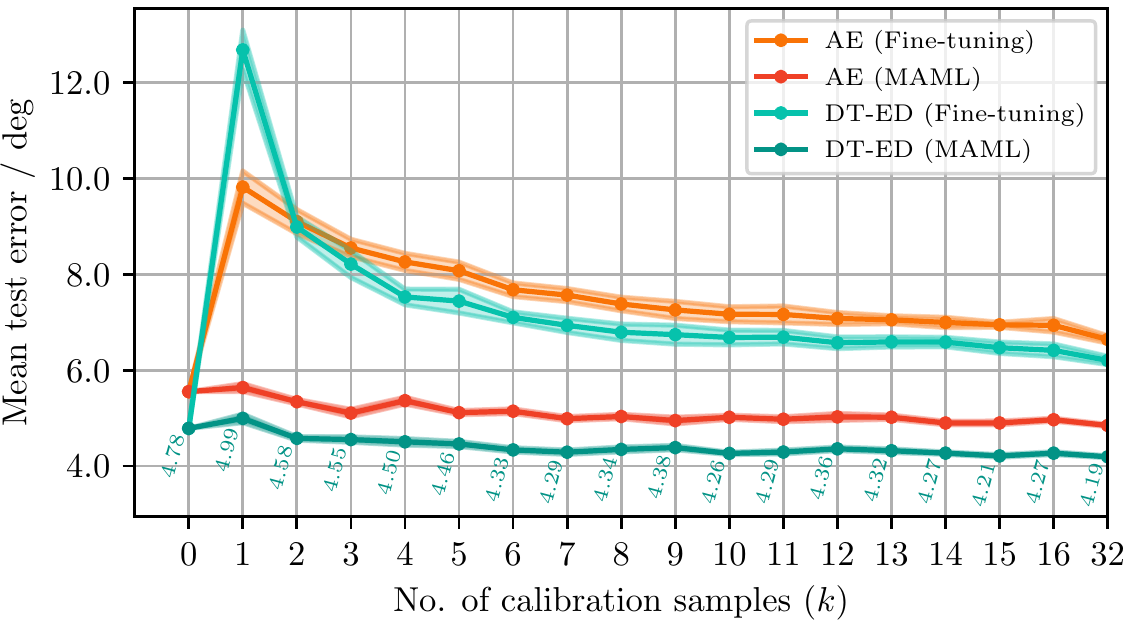}
        \vskip -1mm
        \caption{\vspace{-3mm}}
        \label{fig:maml-vs-nomaml}
    \end{subfigure}
    \hfill
    \begin{subfigure}[b]{0.33\textwidth}
        \includegraphics[width=\columnwidth]{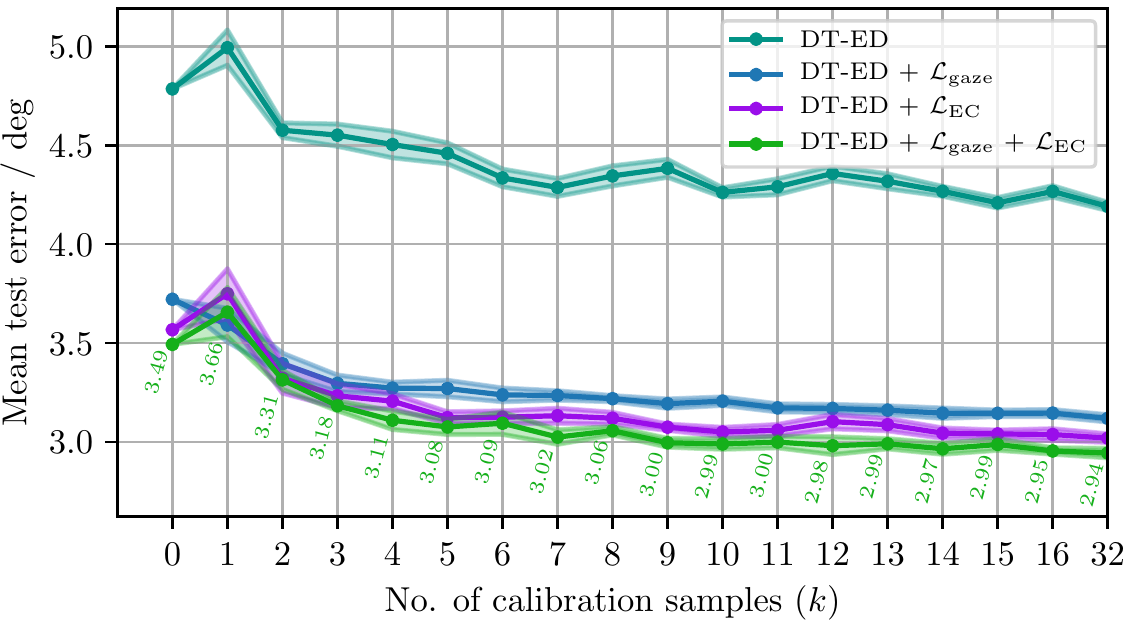}
        \vskip -1mm
        \caption{\vspace{-3mm}}
        \label{fig:impact_of_loss_terms}
    \end{subfigure}
    \begin{subfigure}[b]{0.33\textwidth}
        \includegraphics[width=\columnwidth]{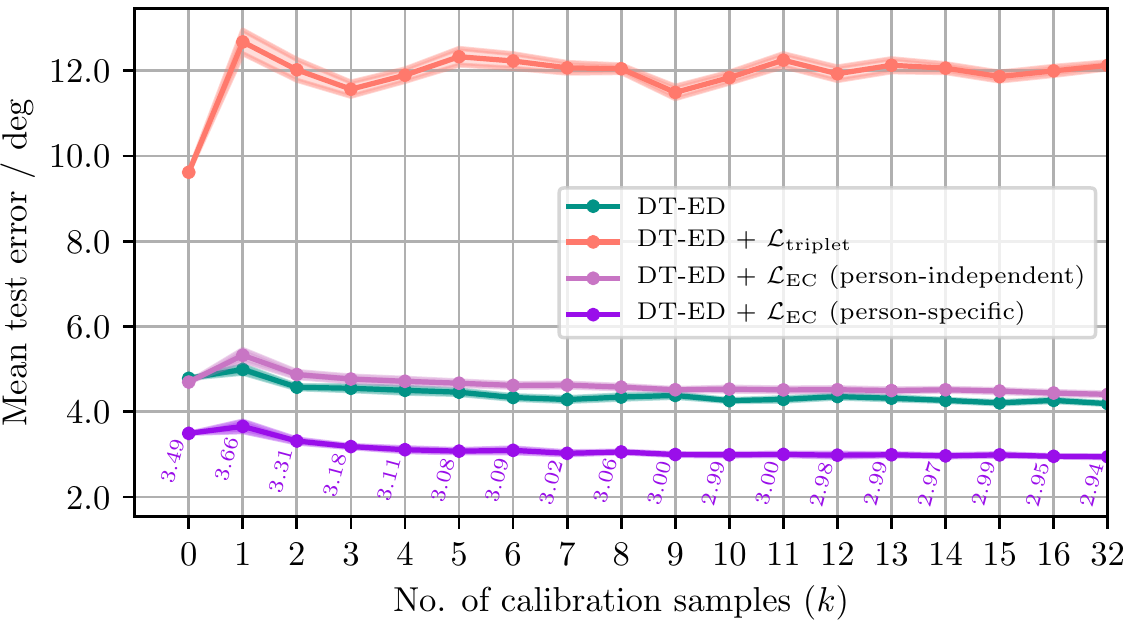}
        \vskip -1mm
        \caption{\vspace{-3mm}}
        \label{fig:impact_of_embedding consistency}
    \end{subfigure}
    \caption{\textbf{Ablation Study on GazeCapture (test):} Impact of (a) learning the few-shot gaze estimator using MAML and using the transforming encoder-decoder for feature learning; (b) different loss terms for training the transforming encoder-decoder; and (c) comparison of the different variants of embedding consistency loss term. 
    }
    \label{fig:ablation}
\end{figure*}

\setcounter{section}{0}
\renewcommand{\thesection}{\Alph{section}}
\section*{\Large Appendix}
\noindent
Due to constraints on the space available in the main paper, we were unable to include all the details there. Here we provide additional implementation details pertaining to our (a) data pre-processing pipeline and (b) the configuration of our DT-ED network. We also show additional results of the ablation study (Section 5.1 in the main paper) on the test partition of the GazeCapture dataset and the performance of \faze for the within MPIIGaze leave-one-person out setting. Finally, we show the sensitivity of \faze to various design configurations.

\section{Implementation Details}
\noindent
We describe further details in how we pre-process the datasets used, and the configuration of the DT-ED architecture.
A reference implementation of both can be found as open-source software at \url{https://github.com/NVlabs/few_shot_gaze}.

\subsection{Data Pre-processing\label{sec:pre-processing}}
\noindent
We employ a normalization procedure based on \cite{Zhang2018ETRA}, which is a revision of \cite{Sugano2014CVPR}, but with a few small changes. We utilize state-of-the-art open-source implementations for face detection\footnote{\tiny\url{https://github.com/cydonia999/Tiny_Faces_in_Tensorflow}} \cite{Hu2017CVPR} and facial landmarks detection\footnote{\tiny\url{https://github.com/jiankangdeng/Face_Detection_Alignment}} \cite{Deng2018FG}, respectively.
 We use the Surrey Face Model \cite{Huber2016} as the reference 3D face model, and select 4 eye corners and 9 nose landmarks as described by the Multi-PIE 68-points markup \cite{Gross2010} for PnP-based \cite{lepetit2009epnp} head pose estimation. This is in contrast to \cite{Sugano2014CVPR,Zhang2018ETRA} which instead use the 4 eye corners and 2 mouth corners. This is motivated by our observation that the mouth corner landmarks are not sufficiently static due to facial expression changes, and that the inherent ambiguity in determining head yaw with very few co-planar landmarks in 3D leads to less reliable head pose estimation.

In our work, we utilize a single image as input which contains both eyes. For this purpose, we select the mean of the 2 inner eye corner landmarks in 3D as the origin of our normalized camera coordinate system.
We use a focal length of $1300mm$ for the normalized camera intrinsic parameters, and a distance of $600mm$ from the face to produce image patches of size $256\times 64$ to use as input for training.

\subsection{Configuration of Disentangling Transforming Encoder-Decoder (DT-ED)}
\noindent
We use the DenseNet architecture to parameterize our encoder-decoder network \cite{Huang2017CVPR}.
We configure the DenseNet with a growth-rate of $32$, 4 dense blocks (each with 4 composite layers), and a compression factor of $1.0$. We neither use dropout nor $1\times 1$ convolutional layers.
We use instance normalization \cite{Ulyanov2016arXiv} and leaky ReLU activation functions (with $\alpha=0.01$) throughout the network as they proved to improve performance for all architectures.

To project CNN features back from latent features $\mathbf{z}$, we apply a fully-connected layer to output values equivalent to $32$ feature maps of width $8$ and height $2$.
The DenseNet decoder that we use to model $\mathcal{D}$ is identical in construction to a usual DenseNet but uses deconvolutional layers (with stride 1) in the place of normal convolutions, and $3\times 3$ deconvolutions (with stride 2) instead of average pooling at the transition layers.
To be faithful to the original implementation, we do not apply bias layers to convolutions in our DenseNet-based DT-ED.
We initialize all layers' weights with MSRA initialization \cite{He2015ICCV}, while biases of the fully-connected layers are initialized with zeros.

\section{Additional Results}
\noindent
We provide additional results of the ablation study on the test partition of the GazeCapture dataset and evaluate the within-dataset performance of \faze on the MPIIGaze dataset.

\subsection{Ablation Study on GazeCapture}
\label{sec:ablation}
\noindent
In the main paper, we provide the results of the ablation study on the MPIIGaze dataset (Fig. 4 in the main paper). Our evaluation setting is a cross-dataset evaluation, where we train on the training partition of the GazeCapture dataset \cite{Krafka2016CVPR} and test on the test partition of the same dataset as well as on MPIIGaze \cite{Zhang2015CVPR}. Here we show additional results for the GazeCapture test partition (Fig. \ref{fig:ablation}).

In Fig.~\ref{fig:maml-vs-nomaml} we observe the same trends for the GazeCapture test dataset that we observed for MPIIGaze. Our proposed DT-ED architecture learns latent representations that are better suited for gaze estimation than those learned by a naive encoder-decoder architecture. Additionally, for few-shot personalization significant gains in accuracy are obtained with meta-learning an adaptable network, as we propose, versus naively fine-tuning a network designed for person-independent gaze estimation (Fine-tuning versus MAML). The latter approach also leads to over-fits at very low $k$. Fig. \ref{fig:impact_of_loss_terms} shows the value of our proposed loss terms of embedding consistency and of computing gaze from the latent representations while training DT-ED, for GazeCapture. Finally, Fig. \ref{fig:impact_of_embedding consistency} shows the consistent improvements obtained for the GazeCapture dataset by preserving inter-person differences versus not doing so.

\begin{figure}
    \centering
    \includegraphics[width=\columnwidth]{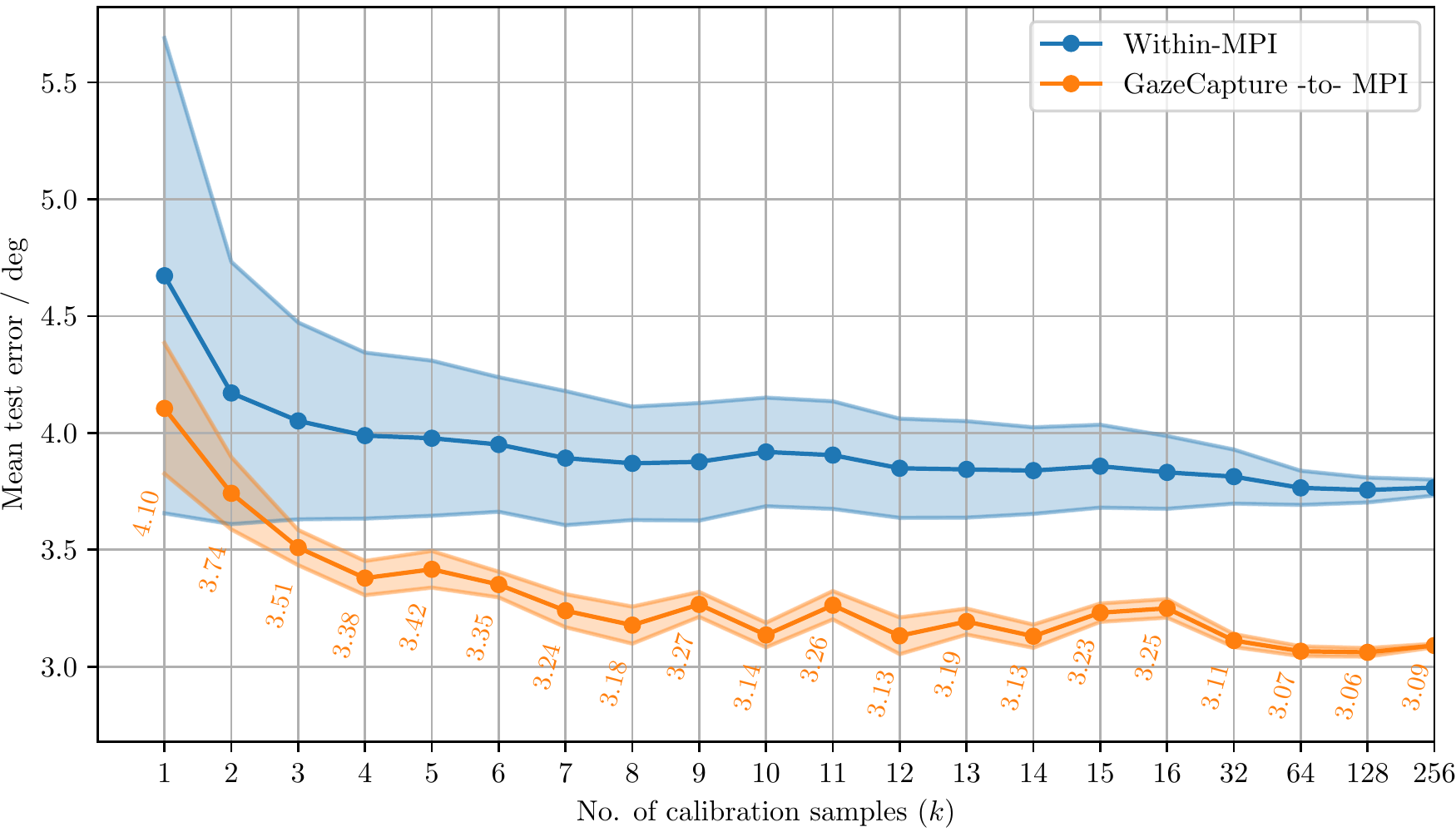}
    \caption{Gaze errors of \faze for within-MPII leave-one-person out training (blue); and training on GazeCapture's training partition and testing one MPIIGaze (orange).}
    \label{fig:within_mpi}
\end{figure}

\subsection{Within-MPIIGaze Performance}

\noindent
So far Liu et al. \cite{Liu2018BMVC} report the best known accuracy of $4.67^\circ$ with $9$ calibration samples on MPIIGaze with their differential network architecture. They use the within MPIIGaze leave-one-subject out evaluation protocol for their experiments. To directly compare against their method, we evaluate the performance of our \faze framework for this experimental protocol (Fig. \ref{fig:within_mpi}). With $9$ calibration samples \faze obtains a gaze error of $3.88$, which is a $17\%$ improvement over Liu et al.'s method. Note, also, that within-MPIIGaze training performs worse than training with GazeCapture (see Fig. 6 in the main paper). This is expected, given the significantly larger diversity of subjects present in the GazeCapture training subset ($993$) versus MPIIGaze ($14$ in a leave-one-out setting), which benefits both DT-ED and MAML. This observation corroborates with similar ones previously made in \cite{Krafka2016CVPR}.

\section{Sensitivity Analysis}
\noindent
We show the influence of various design parameters on the overall performance of our algorithm. These analyses help to determine the parameters' optimal values. 

\subsection{Latent Gaze Code}

\paragraph{Dimension}

\begin{figure}[]
    \vspace*{3cm}
    \centering
    \begin{subfigure}[b]{\columnwidth}
        \includegraphics[width=\columnwidth]{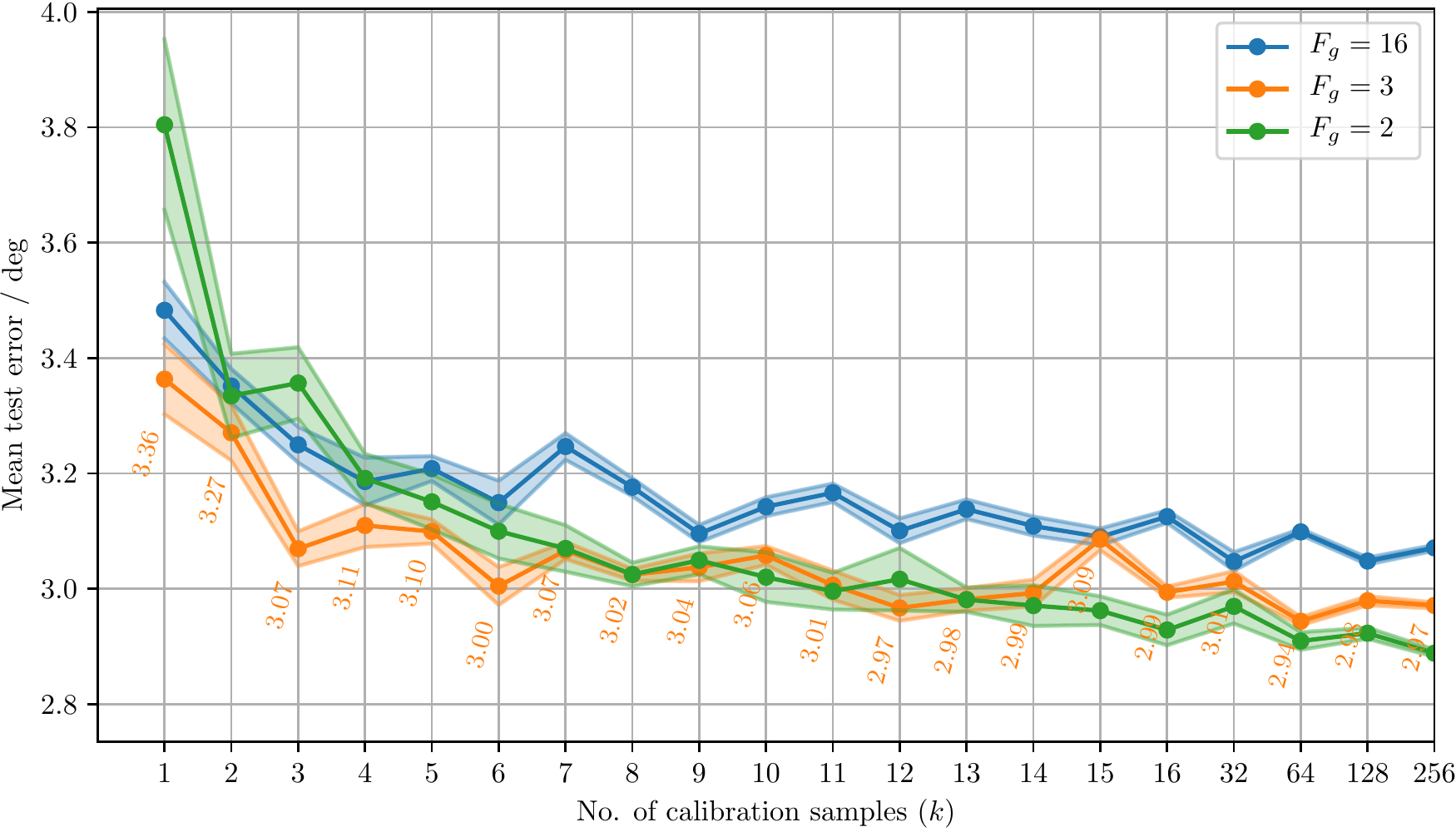}
        \caption{GazeCapture (test)}
    \end{subfigure} \\
    \begin{subfigure}[b]{\columnwidth}
        \includegraphics[width=\columnwidth]{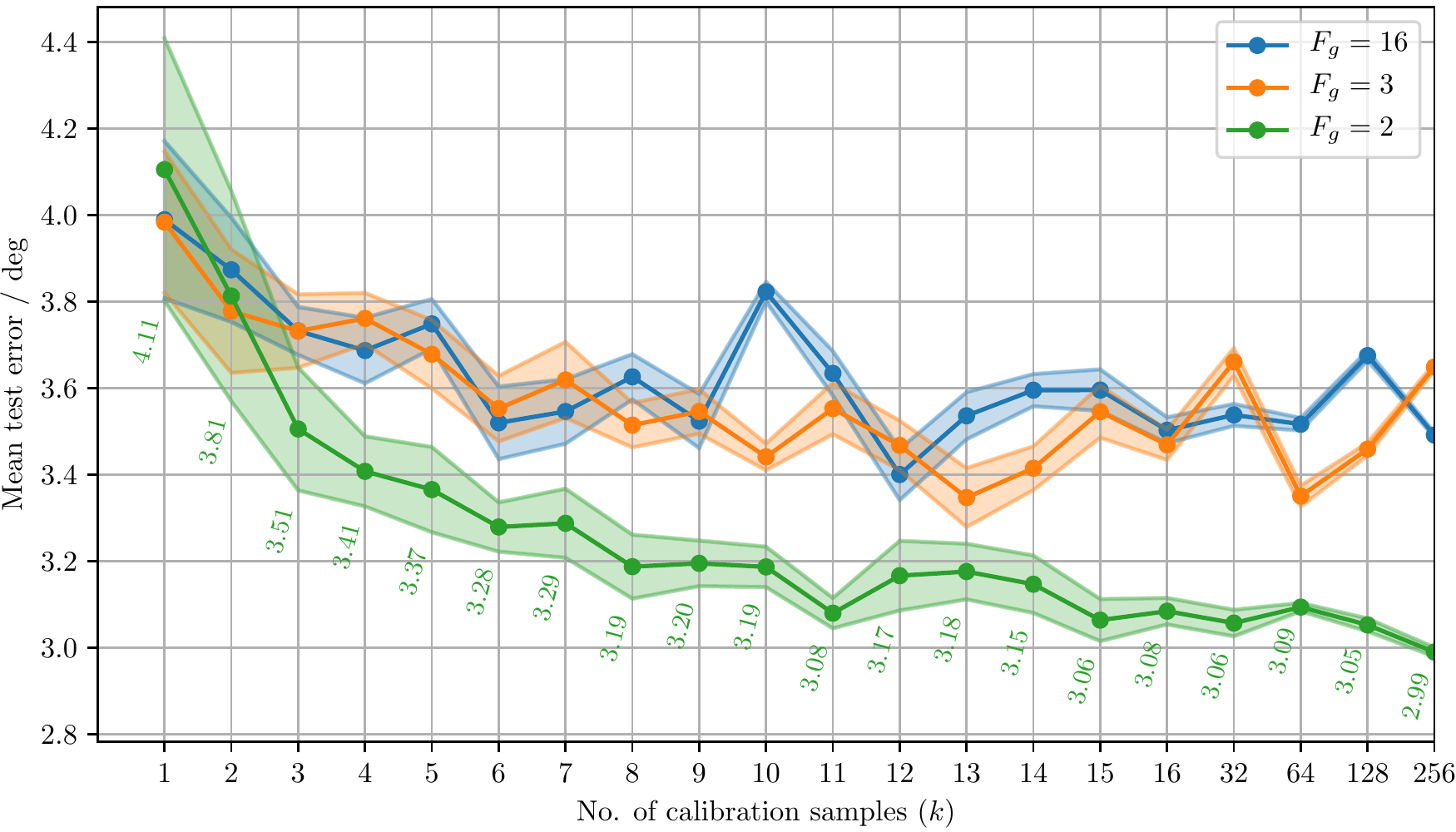}
        \caption{MPIIGaze}
    \end{subfigure}
    \caption{Performance of \faze for different dimensions $F_g$ of the $3\times F_g$-dimensional latent gaze code.}
    \label{fig:gaze_code_dim}
    \vspace*{3cm}
\end{figure}

Our latent gaze code has the dimensions of $3\times F_g$. In order to empirically select the optimal value of $F_g$, we evaluate the performance of \faze for several different values of $F_g = \{16, 3, 2\}$ shown in Fig. \ref{fig:gaze_code_dim}, while keeping the dimensions of the appearance and head pose codes fixed at $64$ and $16$ respectively. Empirically we find $F_g=2$ to be optimal for both datasets and hence select it for our final implementation.

\begin{figure}[]
    \centering
    \begin{subfigure}[b]{\columnwidth}
        \includegraphics[width=\columnwidth]{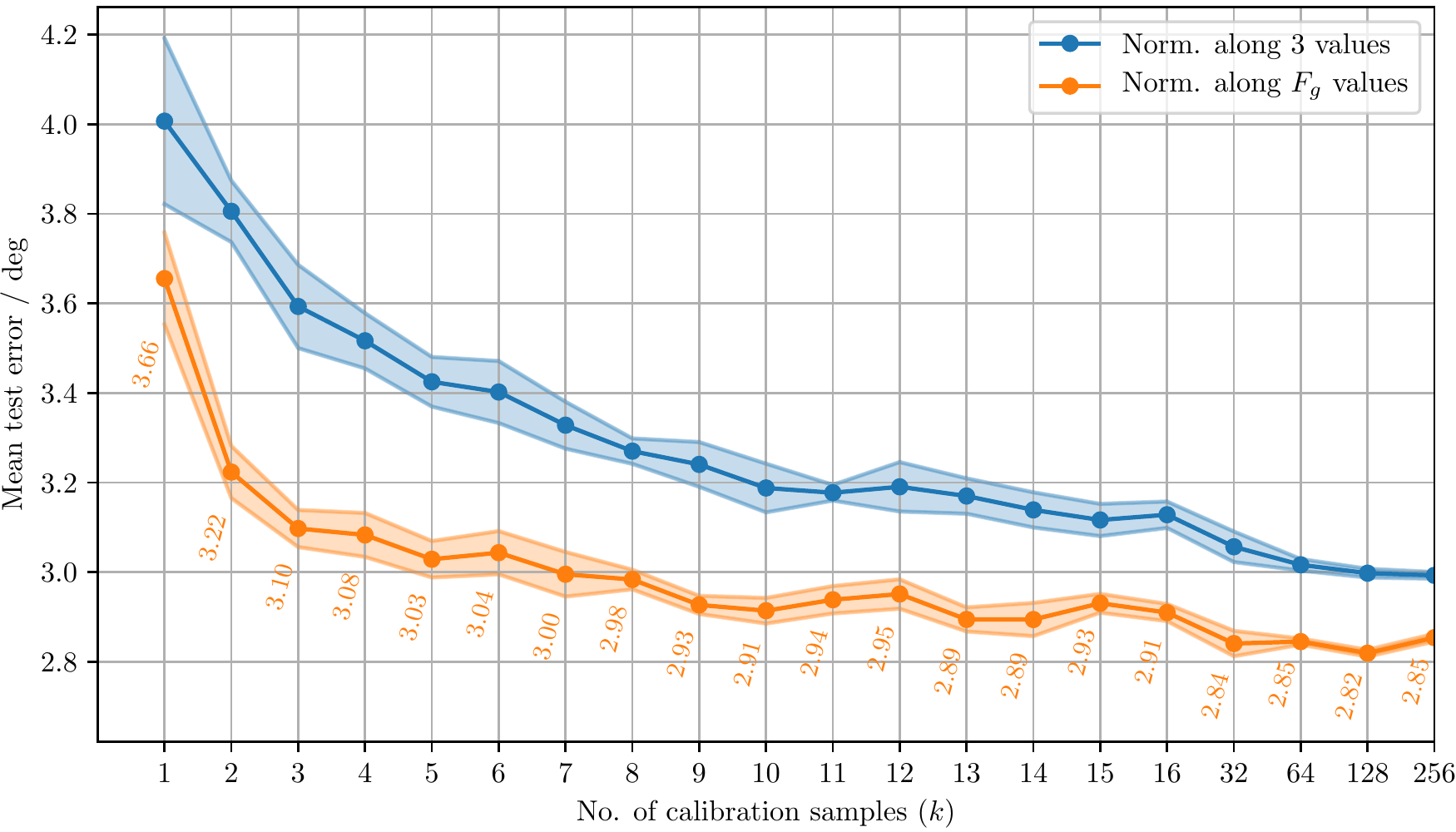}
        \caption{GazeCapture (test)}
    \end{subfigure} \\
    \begin{subfigure}[b]{\columnwidth}
        \includegraphics[width=\columnwidth]{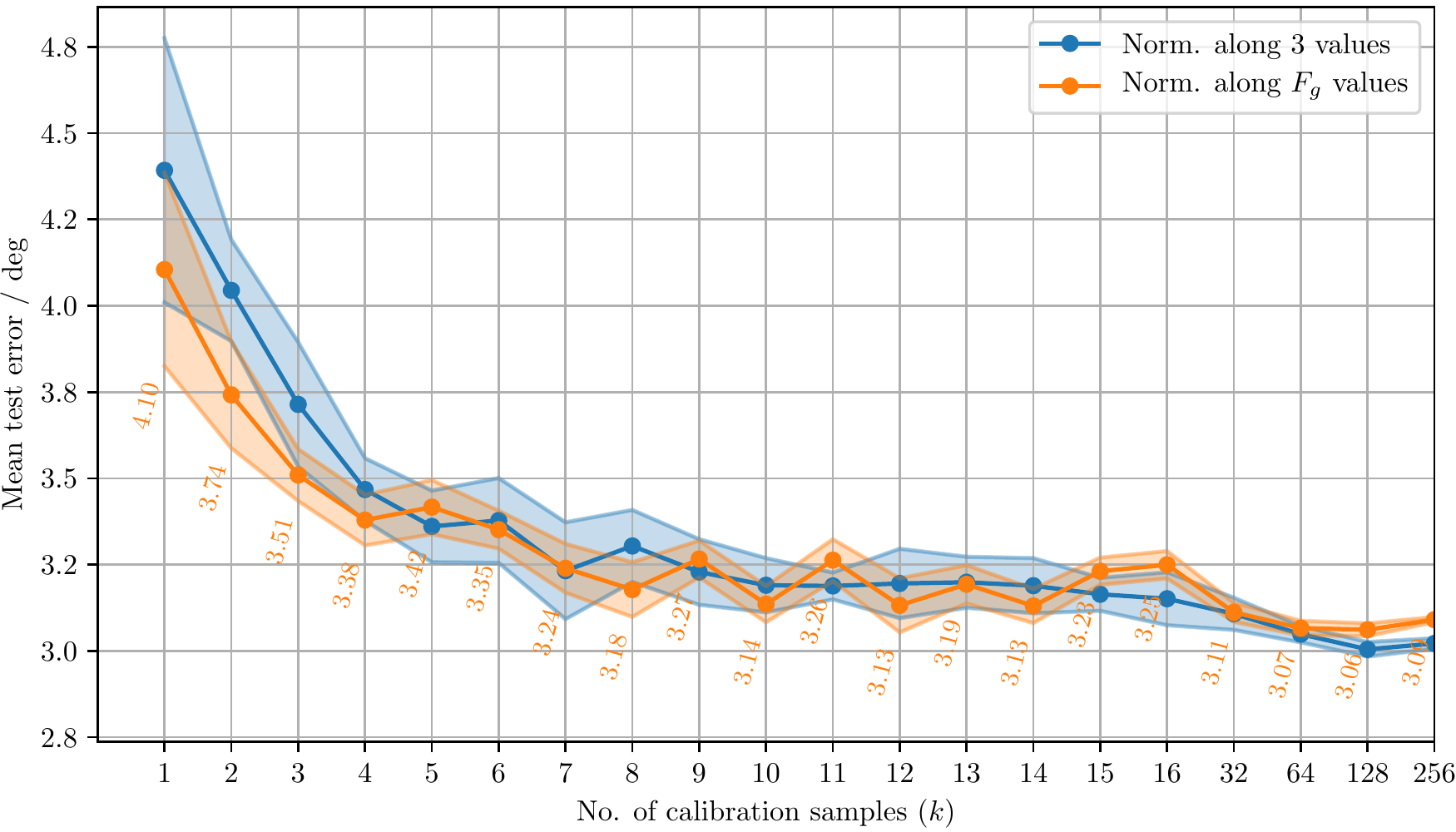}
        \caption{MPIIGaze}
    \end{subfigure}
    \caption{Performance of \faze for normalizing the $3\times F_g$-dimensional gaze code along the $3$ or $F_g$ dimensions, respectively. 
    }
    \label{fig:gaze_code_norm}
\end{figure}

\paragraph{Normalization}

In general we find that is beneficial to normalize our $3\times F_g$-sized latent gaze code to achieve the lowest gaze errors. We experiment with various methods for normalization, which involve computing an $\ell_2$ norm along a particular dimension and dividing all the observed values for that dimension with the norm. We compute norms along the $F_g$ dimension resulting in $3$ norms. Alternatively, one can normalize along the $3$ dimension, resulting in $F_g$ norms. We observe that normalizing along the $F_g$ dimension, produces lower gaze errors for GazeCapture and equivalent ones for MPIIGaze, versus the alternate approach (Fig. \ref{fig:gaze_code_norm}). Hence, we use it for our final implementation.

\end{document}